\PassOptionsToPackage{svgnames,dvipsnames}{xcolor}
\PassOptionsToPackage{capitalize}{cleveref}

\newif\ifarxiv
\arxivtrue

\ifarxiv
 \documentclass[lettersize,twocolumn,twoside]{article}
\else
 \documentclass{CVM}
\fi

\usepackage{color}
\definecolor{TabBlue}{RGB}{78, 121, 167} 
\definecolor{TabLightBlue}{RGB}{160, 203, 232} 
\definecolor{TabOrange}{RGB}{242, 142, 43} 
\definecolor{TabLightOrange}{RGB}{255, 190, 125} 
\definecolor{TabGreen}{RGB}{89, 161, 79} 
\definecolor{TabLightGreen}{RGB}{140, 209, 125} 
\definecolor{TabRed}{RGB}{225, 87, 89} 
\definecolor{TabLightRed}{RGB}{255, 157, 154} 
\definecolor{TabPurple}{RGB}{176, 122, 161} 
\definecolor{TabLightPurple}{RGB}{212, 166, 200} 
\definecolor{TabBrown}{RGB}{157, 118, 96} 
\definecolor{TabLightBrown}{RGB}{215, 181, 166} 
\definecolor{TabPink}{RGB}{211, 114, 149} 
\definecolor{TabLightPink}{RGB}{250, 191, 210} 
\definecolor{TabGray}{RGB}{121, 112, 110} 
\definecolor{TabLightGray}{RGB}{186, 176, 172} 
\definecolor{TabOlive}{RGB}{182, 153, 45} 
\definecolor{TabLightOlive}{RGB}{241, 206, 99} 
\definecolor{TabTeal}{RGB}{73, 152, 148} 
\definecolor{TabLightTeal}{RGB}{134, 188, 182} 

\usepackage[utf8]{inputenc}
\usepackage[T1]{fontenc}
\usepackage{amsmath}
\usepackage{amssymb}
\usepackage{amsfonts}
\DeclareMathSizes{10}{10}{8}{6}

\usepackage{textcomp}
\usepackage{mathtools}
\usepackage{upgreek}
\usepackage{cases}

\usepackage{physics2}
\usephysicsmodule{ab}

\makeatletter
\newcommand\vb{\@ifstar\boldsymbol\mathbf}
\makeatother

\usepackage{doi}
\usepackage{url}
\usepackage{nicefrac} 
\usepackage{microtype} 
\usepackage{graphicx}

\usepackage{tabularray}
\UseTblrLibrary{booktabs,siunitx}
\NewTableCommand{\BB}{\SetCell{font=\bfseries\sffamily}}
\newcommand{\tablefontsize}{\small}
\newcommand{\tablerowsep}{0.5mm}

\usepackage{xspace}
\usepackage{etoolbox}
\usepackage{confnames}

\usepackage{floatrow}
\graphicspath{{./figures}{./figures/mn20}{./figures/iclnuim}{./figures/kitti}}
\DeclareSIUnit{\MB}{\mega\byte}
\DeclareSIUnit{\GB}{\giga\byte}
\DeclareSIUnit[quantity-product=]{\percent}{\char`\%}
\DeclareSIUnit[quantity-product=]{\%}{\char`\%}

\newcommand{\vmu}{\boldsymbol\upmu}
\newcommand{\vphi}{\boldsymbol\upvarphi}

\newcommand{\Vphi}{\boldsymbol\Upphi}

\newcommand{\Vgamma}{\boldsymbol\Upgamma}
\newcommand{\Vsigma}{\boldsymbol\Upsigma}
\newcommand{\Vtheta}{\boldsymbol\Uptheta}

\newcommand{\vbar}{\,|\,}

\newcommand{\DA}{\raisebox{0.4mm}{${\scriptstyle \downarrow}$}}
\newcommand{\UA}{\raisebox{0.4mm}{${\scriptstyle \uparrow}$}}

\DeclareMathOperator*{\argmax}{argmax}

\ifarxiv
 \usepackage{times}
 \usepackage[misc,geometry]{ifsym}
 \NewDocumentCommand{\cor}{}{({\small \Letter})}
 \NewDocumentCommand{\note}{m}{{\footnotesize #1}}

 \usepackage{titlesec}
 \usepackage{titling}
 \titleformat{\section}{\normalfont\large\bfseries}{\thesection}{1em}{}
 \titleformat{\subsection}{\normalfont\bfseries}{\thesubsection}{1em}{}
 \pretitle{\begin{center}\bfseries\Large}
 \posttitle{\par\end{center}\vskip 0.5em}

 \usepackage{fancyhdr}
 \usepackage[twoside,tmargin=2.5cm,bmargin=2.5cm,lmargin=2.5cm,rmargin=2.5cm]{geometry}

 \usepackage[skip=10mm]{caption}
 \hypersetup{colorlinks=true,linkcolor=TabRed,citecolor=TabTeal,urlcolor=TabPink}
\fi

\AtEndPreamble{%
 \usepackage{cleveref}
}

\NewDocumentCommand{\papertitle}{}{Attention-guided reference point shifting for Gaussian-mixture-based partial point set registration}
\NewDocumentCommand{\paperauthors}{}{Mizuki Kikkawa$^{1,*}$, Tatsuya Yatagawa$^{2,*}$\cor{}, Yutaka Ohtake$^{1}$, Hiromasa Suzuki$^{1}$\\ \note{$^{*}$ These authors contributed equally to this study.}}

\ifarxiv
 \title{\papertitle{}}
 \author{\paperauthors{}}
 \date{}
\else
\CVMsetup{%
 type = {Research Article},
 doi = {s41095-0xx-xxxx-x},
 title = {\papertitle{}},
 author = {\paperauthors{}},
 runauthor = {M. Kikkawa et al.},
 abstract = {%
 This study investigates the impact of the invariance of feature vectors for partial-to-partial point set registration under translation and rotation of input point sets, particularly in the realm of techniques based on deep learning and Gaussian mixture models (GMMs). We reveal both theoretical and practical problems associated with such deep-learning-based registration methods using GMMs, with a particular focus on the limitations of DeepGMR, a pioneering study in this line, to the partial-to-partial point set registration. Our primary goal is to uncover the causes behind such methods and propose a comprehensible solution for that. To address this, we introduce an attention-based reference point shifting (ARPS) layer, which robustly identifies a common reference point of two partial point sets, thereby acquiring transformation-invariant features. The ARPS layer employs a well-studied attention module to find a common reference point rather than the overlap region. Owing to this, it significantly enhances the performance of DeepGMR and its recent variant, UGMMReg. Furthermore, these extension models outperform even prior deep learning methods using attention blocks and Transformer to extract the overlap region or common reference points. We believe these findings provide deeper insights into registration methods using deep learning and GMMs. Our source code and datasets are available at \url{https://github.com/tatsy/DGMR-ARPS.git}.
 },
 keywords = {Shape registration; point sets; Gaussian mixture models; deep learning},
 copyright = {The Author(s)},
}
\fi

\begin{document}

\maketitle

\ifarxiv
  \begin{abstract}
    \small
    This study investigates the impact of the invariance of feature vectors for partial-to-partial point set registration under translation and rotation of input point sets, particularly in the realm of techniques based on deep learning and Gaussian mixture models (GMMs). We reveal both theoretical and practical problems associated with such deep-learning-based registration methods using GMMs, with a particular focus on the limitations of DeepGMR, a pioneering study in this line, to the partial-to-partial point set registration. Our primary goal is to uncover the causes behind such methods and propose a comprehensible solution for that. To address this, we introduce an attention-based reference point shifting (ARPS) layer, which robustly identifies a common reference point of two partial point sets, thereby acquiring transformation-invariant features. The ARPS layer employs a well-studied attention module to find a common reference point rather than the overlap region. Owing to this, it significantly enhances the performance of DeepGMR and its recent variant, UGMMReg. Furthermore, these extension models outperform even prior deep learning methods using attention blocks and Transformer to extract the overlap region or common reference points. We believe these findings provide deeper insights into registration methods using deep learning and GMMs. Our source code and datasets are available at \url{https://github.com/tatsy/DGMR-ARPS.git}
  \end{abstract}
\fi

\begin{table}[b!]
  \vspace*{-4mm}
  \small
  \begin{tabular}{p{0.95\linewidth}}
    \toprule \\
  \end{tabular}
  \vspace*{-4mm}
  \noindent \setlength{\tabcolsep}{1pt}
  \begin{tabular}{p{0.05\linewidth}p{0.9\linewidth}}
    $1\quad$ & School of Engineering, The University of Tokyo, 7-3-1, Hongo, Bunkyo-ku, Tokyo, 113-8656, Japan. E-mail: kikkawa.m@nanolab.t.u-tokyo.ac.jp (M. Kikkawa), \{ohtake, suzuki\}@den.t.u-tokyo.ac.jp (Y. Ohtake, H. Suzuki) \\
    $2\quad$ & Graduate School of Social Data Science, Hitotsubashi University, 2-1, Naka, Kunitachi-shi, Tokyo, 186-8601, Japan. E-mail: tatsuya.yatagawa@r.hit-u.ac.jp 
    \vspace{-2mm}
  \end{tabular}
  \vspace{-3mm}
\end{table}

\section{Introduction}
\label{sec:introduction}

\begin{figure}[t!]
  \centering
  \includegraphics[width=\linewidth]{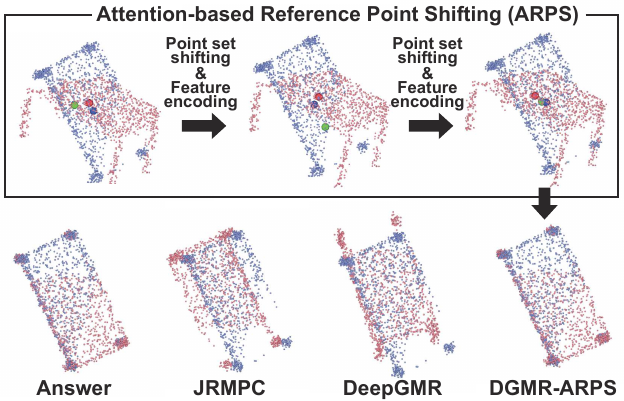}
  \caption{Attention-based reference point shifting (ARPS), proposed in this study, shifts the point sets such that the estimated reference point conforms at the origin, and encodes point positions to obtain point features invariant to rotation and translation. DeepGMR method enhanced by ARPS (DGRM-ARPS) registers point sets more accurately than the traditional method (JRMPC) and deep-learning-based method (DeepGMR) using Gaussian mixture models. In this figure, source and target point sets are represented by red and blue points, and their centroids are by large balls with respective colors, while the green ball represents the origin. ARPS moves these centroids toward the origin.}
  \label{fig:teaser}
\end{figure}

Point set registration is a process of finding rigid transformation(s) to align two or more point sets in different poses and is an essential process for various applications in computer vision, graphics, robotics, and medical image analysis. Point set registration has a closed-form solution if the ideal point-to-point correspondences have been accurately determined. Therefore, previous approaches have aimed to find point-to-point correspondences in various ways, e.g., using distances between points or differences between feature descriptors~\cite{besl1992method,rusinkiewicz2001efficient,segal2010generalized,rusinkiewicz2019symmetric,zhou2016fast,yang2021teaser}. Currently, end-to-end deep learning is a powerful approach to point set registration, in which a neural network is trained to obtain consistent point feature descriptors for two point sets with the same pose~\cite{aoki2019pointnetlk,li2021pointnetlk} or to obtain point feature descriptors to find accurate point-to-point correspondences~\cite{wang2019deep,wang2019prnet,yew2020rpmnet,choy2020deep}.

In contrast, several approaches have associated point sets with probability density functions (PDFs). They use maximum likelihood estimation to find good registration~\cite{myronenko2010point,jian2011robust,eckart2013remseg,eckart2018hgmr,evangelidis2018joint,hirose2021bayesian,zhang2022robust}. Most approaches leverage a Gaussian mixture model (GMM) to obtain the best rigid transformation by iterative numerical optimization, e.g., using expectation-maximization (EM) algorithm~\cite{jian2011robust,eckart2018hgmr,zhang2022robust,eckart2013remseg} or support vector regression~\cite{campbell2015adaptive}. While such PDF-based approaches are known to be more robust to outliers, their main drawback is the slow convergence of the iterative computation required.

Recently, Yuan~et~al.~\cite{yuan2020deepgmr} proposed DeepGMR, the first GMM-based registration method using deep learning. DeepGMR estimates the membership of each point (i.e., how likely each point is to be associated with a Gaussian distribution of the GMM) using a neural network, and represent the input point sets as samples from GMMs. The application of a neural network eliminates a need for time-consuming iterative computation, thereby allowing DeepGMR to significantly cut down the computation time required for traditional GMM-based registration. Furthermore, DeepGMR exhibits greater memory efficiency than point-level registration methods, as the number of mixed distributions is considerably fewer than the number of points.

A problem, however, is that DeepGMR has only been tested on datasets where two point sets have exactly the same point organization but vary in pose. In this case, the centroid (i.e., the average point position) can be used as a reference point to compute consistent point features. However, the performance of DeepGMR decreases significantly for partial point sets with different point organizations because the barycenter of partial point sets do not represent the same reference point in the global object shape. Although the state-of-the-art method, UGMMReg~\cite{huang2022unsupervised}, has attempted to improve DeepGMR using attention blocks~\cite{vaswani2017attention} and a unified GMM~\cite{evangelidis2018joint}, the problem of performance degradation for partial point sets has yet to be solved.

Interestingly, the lack of a known reference point is not a serious problem in non-GMM-based registration methods, and more effort have been made to define consistent point features for partial-to-partial registration~\cite{wang2019prnet,yew2020rpmnet,choy2019fully,choy2020deep}. An exception is the CentroidReg method proposed by Zhao~et~al.~\cite{zhao2021centroidreg}, which performs supervised learning using a neural network to find the centroid of two partial point sets. However, we found that a straightforward combination of DeepGMR with an approach like CentroidReg, where the centroid position is adjusted only once, does not perform well. This implies that the centroid position has a more pronounced impact on the results when using GMM-based methods. In addition, such a point, used as a reference to compute the rotation- and translation-invariant point feature descriptors, does not necessarily have to conform to the centroid of the point sets.

To address this problem, we propose a deep learning technique to find a common reference point in two input point sets. Specifically, our proposed network incorporates attention-based reference point shifting (ARPS), where multi-head attention (MHA) blocks~\cite{vaswani2017attention} are used to enhance point features in the overlap region of two partial point sets. While the idea of applying attention blocks to partial point set registration has been studied for extracting points in the overlap region, our approach leverages the attention blocks to find a single common reference point of two partial point sets. We then shift each point set such that the reference point is at the origin. By repeating the feature extraction with an updated reference point, our network can obtain consistent point features for partial point sets. Despite its simplicity, our approach significantly improves the performance of both the original DeepGMR and its extension, UGMMReg.

In summary, the contributions of this paper are as follows:
\begin{itemize}
  \item This paper reveals theoretical and practical limitations of the previous GMM-based registration method using deep learning, given point sets with different point configurations.
  \item This paper introduces a new point set encoder with reference point shifting, designed to overcome the drawbacks of the previous methods~\cite{yuan2020deepgmr,huang2022unsupervised}.
  \item The proposed method performs better than other GMM-based point set registration methods, including the state-of-the-art deep-learning method~\cite{huang2022unsupervised}.
\end{itemize}

\section{Related Work}
\label{sec:related-work}

\subsection{Traditional point set registration}

When the point-to-point correspondences between two point sets have been accurately determined, aligning the positions of points has a solution in a closed form provided by the singular value decomposition (SVD)~\cite{umeyama1991least}. When the correspondences are unknown, the iterative closest point (ICP)~\cite{besl1992method} is the standard approach, simultaneously solving two problems: finding point-to-point correspondences, and computing a rigid transformation aligning two point sets. While the original algorithm uses spatial proximity of points and associates a point in the source with the nearest point in the target, more robust point-to-point correspondences have been defined by using feature descriptors~\cite{tombari2010shot,choi2012voting,rusu2009fpfh}. After point correspondences are determined, a rigid transformation is computed to minimize the distances between the paired points. While the original ICP minimizes point-to-point distances, other distance metrics, such as point-to-plane~\cite{rusinkiewicz2001efficient}, plane-to-plane~\cite{segal2010generalized}, and symmetric distances~\cite{rusinkiewicz2019symmetric}, have also been exploited. Alongside these point-level methods, the alignment of two or more PDFs has also been investigated. In this approach, each input point set is considered a collection of stochastic samples from a PDF. While most approaches represent the point sets using GMMs~\cite{jian2011robust,eckart2018hgmr,zhang2022robust,eckart2013remseg}, a handful of exceptions have applied the Laplacian distribution~\cite{zhang2022robust} and a more arbitrary distribution defined by variational Bayes~\cite{hirose2021bayesian}.

In addition to developments in ICP, numerous approaches with similar objectives have been introduced in the field of LiDAR odometry. Point set registration methods devised for the LiDAR odometry prioritize computational speed to meet the demands of real-time feedback control of robots. Thus, early approaches often involved relatively straightforward modifications of the original ICP method. For example, TriICP~\cite{chetverikov2002trimmed} employs a trimmed squared distance, and the normal distribution transform~\cite{biber2003normal}, similar to Gaussian mixture registration, aligns two normal distributions associated with point sets. Other approaches dedicated to LiDAR odometry employ not only scan-to-scan alignment, but also scan-to-map alignment to improve alignment accuracy. Furthermore, several algorithms focus on the distinctive parts of geometries to simplify the alignment calculation. For example, LOAM~\cite{zhang2014loam} and its variants~\cite{zhang2016low,wang2021f} utilize sharp features and planar surface patches to accelerate point set registration. Recently, KISS-ICP~\cite{vizzo2023kiss} provides a solution to align multiple point sets obtained by LiDARs, which differ in terms of field of views, resolutions, etc. For related studies concerning LiDAR odometry, including those using deep learning, we refer the readers to a comprehensive survey~\cite{lee2024lidar}.

\subsection{Point set registration using deep learning}

Recent deep-learning-based approaches solve point set registration problems in an end-to-end manner~\cite{aoki2019pointnetlk,wang2019deep}. Most approaches compute feature vectors using a convolutional neural network (CNN) on the point sets, such as PointNet~\cite{qi2017pointnet,qi2017pointnet2} and dynamic graph CNN (DGCNN)~\cite{wang2019dynamic}. Based on the point correspondences defined using the learned features, a rigid transformation is computed by a differentiable registration module, such as a Lucas--Kanade layer~\cite{aoki2019pointnetlk,li2021pointnetlk} or a differentiable SVD layer~\cite{wang2019deep}. Following these early studies, recent approaches have achieved partial-to-partial registration by extracting the overlap region of two input point sets, often applying attention and Transformer blocks~\cite{vaswani2017attention} to extract points in the overlap regions~\cite{wang2019deep,choy2020deep,lee2021deep,yew2022regtr}. Several others, like RANSAC, exclude points in the non-overlapping regions as outliers. For example, robust point matching (RPM)~\cite{yew2020rpmnet} and the loss function to measure the extent of global alignment~\cite{shen2022reliable} are employed. Furthermore, the application of graph neural networks to point set registration is a recent trend~\cite{min2021distinctiveness,fu2021robust}. Although the GMM-based point set registration using deep learning~\cite{yuan2020deepgmr,huang2022unsupervised} has also been proposed recently, using it for partial point sets has yet to be sufficiently investigated.

\subsection{Rotation-invariant feature descriptors}

PointNet~\cite{qi2017pointnet} and PointNet++~\cite{qi2017pointnet2}, the pioneering studies in learning permutation-invariant feature extraction using a deep neural network, process each point individually using a multi-layer perceptron (MLP) and aggregate features globally or locally using a pooling operation. Later studies improved the point-wise feature extraction, e.g., by feeding handcrafted features~\cite{deng2018ppfnet,chen2019clusternet,gojcic2019perfect} to a neural network and enhanced feature aggregation~\cite{wang2019dynamic,xu2018spidercnn,li2018pointcnn}. Furthermore, several previous studies have proposed techniques to achieve rotation invariance of feature descriptors. Some have used the directions of point normals~\cite{khoury2017learning,ao2021spinnet}, others have relied on arrangements of local point subsets~\cite{gojcic2019perfect,spezialetti2019learning}, and yet others have utilized the centroid of the global shape~\cite{chen2019clusternet,zhang2019rotation,yu2020deep}. Although the last, leveraging the centroid, has implicitly been assumed by DeepGMR~\cite{yuan2020deepgmr}, the centroids of partial point sets do not represent the same position in the global shape. For partial point set registration, Zhao~et~al.~\cite{zhao2021centroidreg} performed supervised learning of the centroid position to find rotation-invariant point features. However, the impact of their method for GMM-based deep learning methods has yet to be investigated.

\section{Background}
\label{sec:background}

\begin{figure}[t!]
  \centering
  \includegraphics[width=\linewidth]{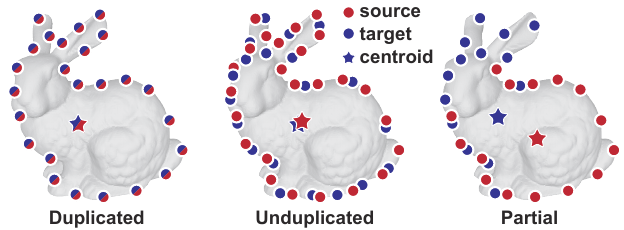}
  \caption{We categorize input point sets based on the point arrangement after registration. \textit{Duplicated} point sets can be aligned perfectly such that all the source points coincide with the target points after the registration. \textit{Unduplicated} point sets both represent the global shape of an object but do not have the same point configuration, and their points are not aligned perfectly by registration. \textit{Partial} point sets represent different parts of the object's geometry, and only parts of their shapes overlap even after registration.}
  \label{fig:pointset-types}
\end{figure}

This section reviews the original DeepGMR before introducing our method. In the following, we refer differently to a pair of source and target point sets with different point configurations, as illustrated in \cref{fig:pointset-types}. When the positions of corresponding points in both sets can be brought into conformation by registration, we refer to the pair as \textit{duplicated} point sets. Otherwise, we refer to the pair as \textit{unduplicated} point sets. Furthermore, when source and target point sets of an unduplicated pair represent different parts of the global shape, we refer to the pair as \textit{partial} point sets.

DeepGMR is relatively robust in the face of point sets with substantially different poses, while other deep-learning-based approaches work only for point sets with a comparatively small difference in orientation and position. However, we have found that DeepGMR is only sure to succeed for duplicated point sets. Unfortunately, its performance worsens significantly for partial point sets. We consider the reason in this section.

\subsection{Point set registration using Gaussian mixtures}

Let $\hat{\mathcal{X}} \subset \mathbb{R}^3$ and $\mathcal{X} \subset \mathbb{R}^3$ be the source and target point sets that we wish to align. For simplicity, we assume that both point sets include the same number of points, $N$.
\begin{subequations}
  \begin{align}
    \hat{\mathcal{X}} &= \{ \hat{\vb{x}}_i \in \mathbb{R}^3 : i = 1, \ldots, N \}, \\
    \mathcal{X}       &= \{ \vb{x}_i \in \mathbb{R}^3 : i = 1, \ldots, N \},
  \end{align}
\end{subequations}
where $\hat{\vb{x}}_i$ and $\vb{x}_i$ with the same index $i$ do not necessarily correspond to each other. We align $\hat{\mathcal{X}}$ to $\mathcal{X}$ by applying a rigid transformation $T \in \mathrm{SE}(3)$ to $\hat{\mathcal{X}}$. Specifically, $T(\hat{\vb{x}}) = \vb{R} \hat{\vb{x}} + \vb{t}$, where $\vb{R} \in \mathrm{SO}(3)$ and $\vb{t} \in \mathbb{R}^3$ are a rotation matrix and a translation vector, respectively.

GMM-based registration~\cite{jian2011robust,eckart2018hgmr} assumes that a point set consists of stochastic samples independently and identically drawn from a GMM. Let $\hat{\Vtheta}$ and $\Vtheta$ be parameter sets for the GMMs of $\hat{\mathcal{X}}$ and $\mathcal{X}$, respectively. Then, a GMM $P$ represents a multimodal probability distribution using convex combinations of Gaussian distributions.
\begin{equation}
  P(\vb{x} \vbar \Vtheta) = \sum_{j=1}^J \pi_j G(\vb{x} \vbar \vmu_j, \Vsigma_j),
\end{equation}
where $\sum_j \pi_j = 1$, the parameter set $\Vtheta$ consists of $J$ triplets $(\pi_j, \vmu_j, \Vsigma_j)$ corresponding to mixture weight, mean, and covariance matrix for the $j$th Gaussian distribution; $G(\vb{x} \vbar \vmu, \Vsigma)$ is a Gaussian distribution with mean $\vmu$ and covariance $\Vsigma$. Rather than minimizing point-to-point distances, GMM-based methods employ maximum likelihood estimation (MLE) to obtain the rigid transformation $T$.
\begin{equation}
  \begin{aligned}[b]
    T^{*}
     & = \argmax_{T \in \mathrm{SE}(3)} \mathbb{E}_{\hat{\vb{x}} \in \hat{\mathcal{X}}} \ab[ \log P(T(\hat{\vb{x}}) \vbar \Vtheta) ] \\
     & = \argmax_{T \in \mathrm{SE}(3)} \sum_{i=1}^N \log \sum_{j=1}^J \pi_j G(\hat{\vb{x}}_i \vbar \vmu_j, \Vsigma_j).
  \end{aligned}
\end{equation}
The EM algorithm is often employed to solve this maximization problem. Given point sets $\hat{\mathcal{X}}$ and $\mathcal{X}$, the algorithm alternates the following two steps~\cite{eckart2018hgmr,yuan2020deepgmr}.
\begin{subequations}
  \begin{alignat}{2}
    \text{\bf E step:} & \quad \Vtheta^{*} = \argmax_{\Vtheta} \mathbb{E}_{\vb{x} \in \mathcal{X}} \ab[ P(\vb{x} \vbar \Vtheta) ], \label{eq:e-step}                                      \\
    \text{\bf M step:} & \quad T^{*} = \argmax_{T \in \mathrm{SE}(3)} \mathbb{E}_{\hat{\vb{x}} \in \hat{\mathcal{X}}} \ab[ \log P(T(\hat{\vb{x}}) \vbar \Vtheta^{*}) ]. \label{eq:m-step}
  \end{alignat}
\end{subequations}

\begin{table*}[t!]
  \centering
  \caption{DeepGMR for point sets with different configurations}
  \label{tab:deepgmr-pretest}
  {\tablefontsize
    \begin{tblr}{%
        colspec={%
            Q[l]
            Q[c,si={table-format=2.4,round-mode=places,round-precision=4}]
            Q[c,si={table-format=1.4,round-mode=places,round-precision=4}]
            Q[c,si={table-format=1.4,round-mode=places,round-precision=4}]
            Q[c,si={table-format=1.4,round-mode=places,round-precision=4}]
            Q[c,si={table-format=1.4,round-mode=places,round-precision=4}]
          },%
        rowsep=\tablerowsep,
        colsep=5mm,
        hline{2}={1}{rightpos=-0.2},%
        hline{2}={2-3}{rightpos=-0.2,leftpos=-0.2,endpos},%
        hline{2}={4-7}{leftpos=-0.2,endpos},%
      }
      \toprule
      Input & {{{MRE \DA}}} & {{{MTE \DA}}} & {{{Recall \UA}}} & {{{RRE \DA}}} & {{{RTE \DA}}} \\
      D-XYZ & 1.743738      & 0.008837      & 0.998782         & 1.723405      & 0.008731      \\
      U-XYZ & 7.464414      & 0.049038      & 0.870942         & 4.426355      & 0.033688      \\
      P-XYZ & 19.232632     & 0.219192      & 0.395698         & 8.267095      & 0.147870      \\
      D-RRI & 1.622372      & 0.008226      & 1.000000         & 1.622372      & 0.008226      \\
      U-RRI & 27.106122     & 0.122758      & 0.625406         & 7.052702      & 0.043854      \\
      P-RRI & 66.683691     & 0.362073      & 0.129464         & 9.853736      & 0.119397      \\
      \bottomrule
    \end{tblr}
  }
\end{table*}

\subsection{DeepGMR}
\label{ssec:deepgmr}

DeepGMR solves the two optimization problems in \cref{eq:e-step,eq:m-step} using deep learning. Specifically, a neural network is used to solve the E step to avoid the iterative optimization alternating the E and M steps. We hereinafter refer to the network as a backbone network. As in the E step of the standard EM algorithm, the backbone network infers the membership $\gamma_{ij}$, which represents how likely the point $\vb{x}_i$ is associated with $j$th Gaussian distribution. In DeepGMR, each Gaussian distribution is represented as an isotropic distribution with $\Vsigma_j = \sigma_j^2 \vb{I}$, where $\vb{I} \in \mathbb{R}^{3 \times 3}$ is an identity matrix. Thus, the parameter triplet $(\pi_j, \vmu_j, \sigma_j^2)$ to represent a Gaussian distribution is obtained as follows:
\begin{subequations}
  \begin{align}
    \pi_j      &= \frac{1}{N} \sum_{i=1}^N \gamma_{ij}, \\
    \vmu_j     &= \frac{1}{N\pi_j} \sum_{i=1}^N \gamma_{ij} \vb{x}_i, \\
    \sigma_j^2 &= \frac{1}{3 N\pi_j} \sum_{i=1}^N {(\vb{x}_i - \vmu_j)}^\top (\vb{x} - \vmu_j).
  \end{align}
\end{subequations}
Using this representation, we can rewrite the M step to obtain the best rigid transformation $T^{*}$ as
\begin{equation}
  T^{*} = \underset{T \in \mathrm{SE}(3)}{\mathrm{argmin}} \sum_{j=1}^{J} \frac{\hat{\pi}_j}{\sigma_j^2} \left\lVert T(\hat{\vmu}_j) - \vmu_j \right\rVert^2,
  \label{eq:modified-m-step}
\end{equation}
where $\| \cdot \|$ is the Euclidean norm. \Cref{eq:modified-m-step} is a weighted form of the ordinary point-to-point set registration problem, and can be solved using SVD~\cite{umeyama1991least}. For more details, refer to ~\cite{yuan2020deepgmr} and its supplementary document.

\subsection{Limitations of DeepGMR}
\label{ssec:problem-deepgmr}

DeepGMR inputs the rigorously rotation-invariant (RRI) feature~\cite{chen2019clusternet} to the backbone network. From the handcrafted RRI features, the backbone network can obtain learned rotationally invariant features more easily. An RRI feature consists of four types of scalars: $r_i$, $r_{ik}$, $\theta_{ik}$, and $\phi_{ik}$ computed for $\vb{x}_i$ and its $M$ nearest neighbors $\vb{x}_{i1}, \ldots, \vb{x}_{iM}$. Therefore, each RRI feature is a $4M$-dimensional vector ($r_i$ is repeated redundantly for all neighbors).

RRI features for a point set $\{ \vb{x}_i \}$, which we denote as $\Vphi$, do not change when the point set is rotated around an arbitrary rotation center. Therefore, we can obtain consistent features for two or more point sets, provided that we know the rotation center in advance. Consider a case where $\hat{\mathcal{X}}$ and $\mathcal{X}$ are a duplicated point set pair (i.e., $\hat{\mathcal{X}} = \{ T(\vb{x}_i) : \vb{x}_i \in \mathcal{X} \}$). In this case, the centroids $\hat{\vb{c}}$ and $\vb{c}$ of the two point sets can be considered as a rotation center (which is also a reference point), and all four scalars of $\hat{\vb{x}}_i$ will be exactly the same as $\vb{x}_i$. For example, the following equation holds for $r_i$.
\begin{equation}
  \hat{r}_i = \| \hat{\vb{x}}_i - \hat{\vb{c}} \| = \| \vb{R} (\vb{x}_i - \vb{c}) \| = \| \vb{x}_i - \vb{c} \| = r_i.
\end{equation}
Thus, the RRI feature of each $\hat{\vb{x}}_i$ perfectly matches that of $\vb{x}_i$ for duplicated point sets. Without loss of generality, we can translate two point sets such that the positions of $\hat{\vb{c}}$ and $\vb{c}$ conform with the origin and use the origin as a reference point. Owing to the consistency of input RRI features, DeepGMR can obtain consistent GMMs for point set registration.

Since DeepGMR transforms the point features $\Vphi$ to the memberships $\Vgamma = [ \gamma_{ij} ]$ using the backbone network, the memberships $\hat{\gamma_{ij}}$ and $\gamma_{ij}$ will be consistent as well. Then, $(\hat{\pi}_j, \hat{\vmu}_j, \hat{\Vsigma}_j)$ and $(\pi_j, \vmu_j, \Vsigma_j)$ are consistent for each $j$. More importantly, the consistency of the GMM parameters holds when the point-to-point correspondences are unknown because the computation of the parameters is invariant with respect to reordering the points. Due to the independence of the correct correspondences, DeepGMR performs well for duplicated point sets.

However, the centroids of unduplicated and partial point sets do not represent the same reference point. \Cref{fig:pointset-types} illustrates this problem, where red and blue circles indicate points in source and target point sets, respectively, and a star indicates their centroids. As can be seen, the centroids of unduplicated and partial point sets do not lie at the same position in the global shape, while those for the duplicated point sets coincide. Therefore, the point features are no longer invariant under rigid transformation when the centroid of each point set is used as a reference point.

\Cref{tab:deepgmr-pretest} shows the results of a preliminary experiment assessing the performance of DeepGMR with changes in input features (XYZ coordinates or RRI features) and point organization (\textbf{D}=duplicated, \textbf{U}=unduplicated, \textbf{P}=partial). This experiment was conducted using the ModelNet40 dataset~\cite{wu2015shapenets}; the performance was evaluated by mean rotation/translation errors (MRE/MTE), recall, and average errors for \textit{recalled} (i.e., inlier) rotation/translation (RRE/RTE). The rotation errors are computed as cosine distance between two rotation matrices, and the translation errors are computed as Euclidean distance between two translation vectors~\cite{choy2020deep}. To compute the recall, RRE, and RTE, we regard a rigid transformation as an inlier when the rotation error is less than \ang{15} and the translation error is less than 0.2. Note that the size of each point set is normalized to fit a cube ${[-1, 1]}^3$ by keeping an aspect ratio. The detailed experimental setup is described later in \cref{sec:experiments}.

As shown in \cref{tab:deepgmr-pretest}, the performance of DeepGMR decreases significantly when the input point sets are partial. Furthermore, the performance for partial point sets using RRI features is limited compared to when using XYZ coordinates as inputs. This result suggests that the feature descriptors encoded from XYZ coordinates by a neural network are more effective than handcrafted RRI features for partial-to-partial registration. In summary, the preliminary experiment shows:
\begin{enumerate}
  \item DeepGMR performs well when input point sets are translated so that a reference point within them coincides with the origin;
  \item XYZ coordinates are a better choice as inputs to the network unless the input point sets are duplicated.
\end{enumerate}
Based on these observations, we design our proposed network to find a common reference point for two input point sets using an architecture similar to Transformer~\cite{vaswani2017attention} and input XYZ coordinates of points to the network rather than RRI features.

\section{Attention-Guided Reference Point Shifting}
\label{sec:arps-layer}

\begin{figure}[t!]
  \centering
  \hspace{8mm}
  \includegraphics[width=0.9\linewidth]{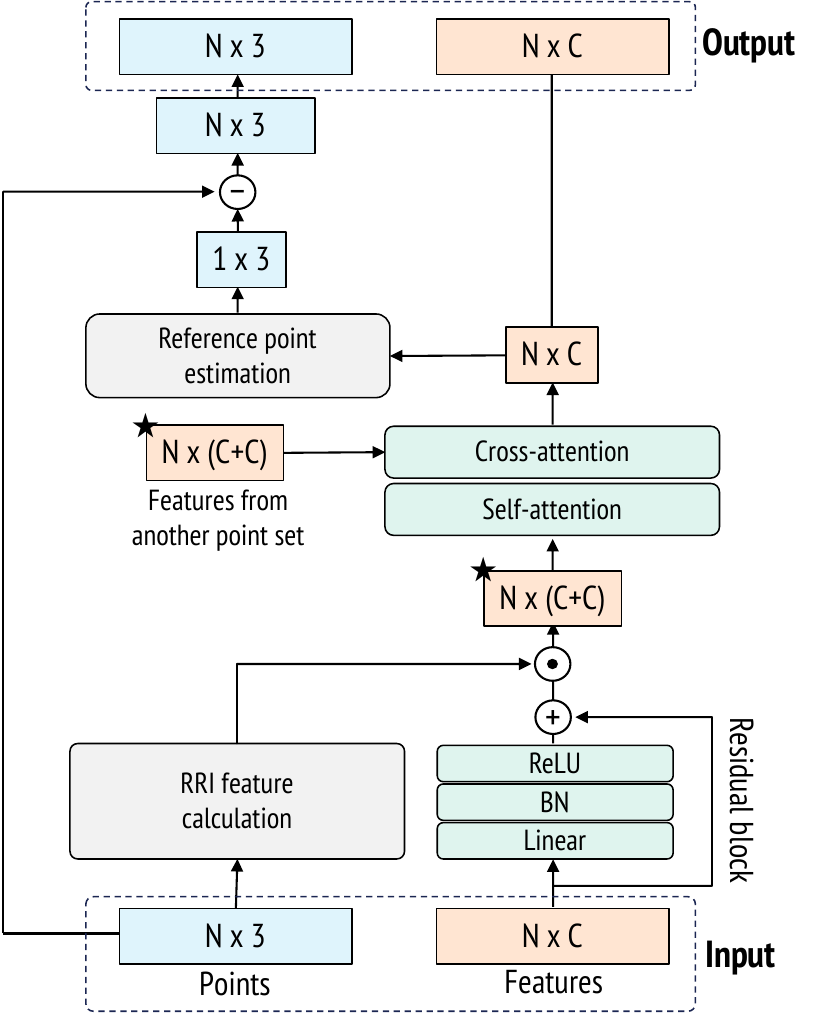}
  \caption{Detailed construction of the ARPS layer. One of the inputs for the cross-attention blocks is obtained from the other point set. The features with a $\star$ correspond to each other.}
  \label{fig:arps-layer}
\end{figure}

Based on the previous discussion, we install ARPS layers into the neural network to find a common reference point for two partial point sets. We show the detailed configuration of a single ARPS layer in \cref{fig:arps-layer}. The input for an ARPS layer comprises pairs of position $\vb{x}_i \in \mathbb{R}^3$ and feature $\vb{f}_i \in \mathbb{R}^C$ for points, where $C$ is the dimension of a feature vector. An ARPS operation begins by encoding the XYZ coordinates of points using an MLP, and we obtain a feature descriptor $\vphi_i \in \mathbb{R}^{C}$ for each point, which is concatenated with the feature $\vb{f}_i$ from the previous layer. This process is performed for both source and target point sets to give $\hat{\mathcal{F}} = \lbrace (\hat{\vb{f}}_i, \hat{\vphi}_i) \rbrace_{i=1}^N$ and $\mathcal{F} = \lbrace (\vb{f}_i, \vphi_i) \rbrace_{i=1}^N$.

\subsection{ARPS layer}

To find a common reference point, we employ MHA blocks to enhance the features of points in the overlap region. An MHA block consists of several attention blocks, each of which computes its output as
\begin{subequations}
  \begin{align}
    \mathrm{MHA}(\vb{X},\!\vb{Y},\!\vb{Z}) \! &=\! \mathrm{FFN}(\mathrm{concat}(\vb{H}_1, \ldots, \vb{H}_h) \vb{W}^O), \\
    \vb{H}_h                                  &= \mathrm{Attn}(\vb{X} \vb{W}_h^Q, \vb{Y} \vb{W}_h^K, \vb{Z} \vb{W}_h^V), \\
    \mathrm{Attn}(\vb{Q}, \vb{K}, \vb{V})     &= \mathrm{softmax} \left( \frac{\vb{Q} \vb{K}^T}{\sqrt{d_k}} \right) \vb{V},
  \end{align}
\end{subequations}
where $\vb{X}, \vb{Y}, \vb{Z} \in \mathbb{R}^{N\times d_m}$ are the inputs to the MHA, $\vb{W}_h^Q, \vb{W}_h^K \in \mathbb{R}^{d_m \times d_k}$ and $\vb{W}_h^V \in \mathbb{R}^{d_m\times d_v}$ are parameter matrices of the $h$th attention block, and $\mathrm{FFN}(\: \cdot \:)$ is a position-wise feed-forward network. In the ARPS layer, the set of features $\hat{\mathcal{F}}$ and $\mathcal{F}$ are enhanced by self- and cross-attentions defined respectively by two Siamese MHAs.
\begin{subnumcases}{\hspace{-1.8em}}
  \hat{\mathcal{F}}^{'} = \mathrm{MHA}_{\text{self}} (\hat{\mathcal{F}}, \hat{\mathcal{F}}, \hat{\mathcal{F}}), \\
  \mathcal{F}^{'} = \mathrm{MHA}_{\text{self}} (\mathcal{F}, \mathcal{F}, \mathcal{F}),
\end{subnumcases}
\begin{subnumcases}{}
  \hat{\mathcal{F}}^{''} = \mathrm{MHA}_{\text{cross}} (\hat{\mathcal{F}}^{'}, \mathcal{F}^{'}, \mathcal{F}^{'}), \\
  \mathcal{F}^{''} = \mathrm{MHA}_{\text{cross}} (\mathcal{F}^{'}, \hat{\mathcal{F}}^{'}, \hat{\mathcal{F}}^{'}).
\end{subnumcases}
Then, we extract feature subsets from $\hat{\mathcal{F}}''$ and $\mathcal{F}''$ by collecting indices $\mathcal{I}_H$ of the top $H$ features in the order of descending magnitude of feature norms.
\begin{subequations}
  \begin{align}
    \hat{\mathcal{I}}_H &= \mathrm{top}_H(\| \hat{\vb{f}}''_1 \|, \ldots, \| \hat{\vb{f}}''_N \|), \\
    \mathcal{I}_H       &= \mathrm{top}_H(\| \vb{f}''_1 \|, \ldots, \| \vb{f}''_N \|),
  \end{align}
\end{subequations}
where $\hat{\vb{f}}''_i$ and $\vb{f}''_i$ are the $i$th feature vectors in $\hat{\mathcal{F}}''$ and $\mathcal{F}''$, respectively. Now, we can assume that the index sets $\hat{\mathcal{I}}_H$ and $\mathcal{I}_H$ represent the points in the overlap region.

\begin{figure*}[t!]
  \centering
  \includegraphics[width=\linewidth]{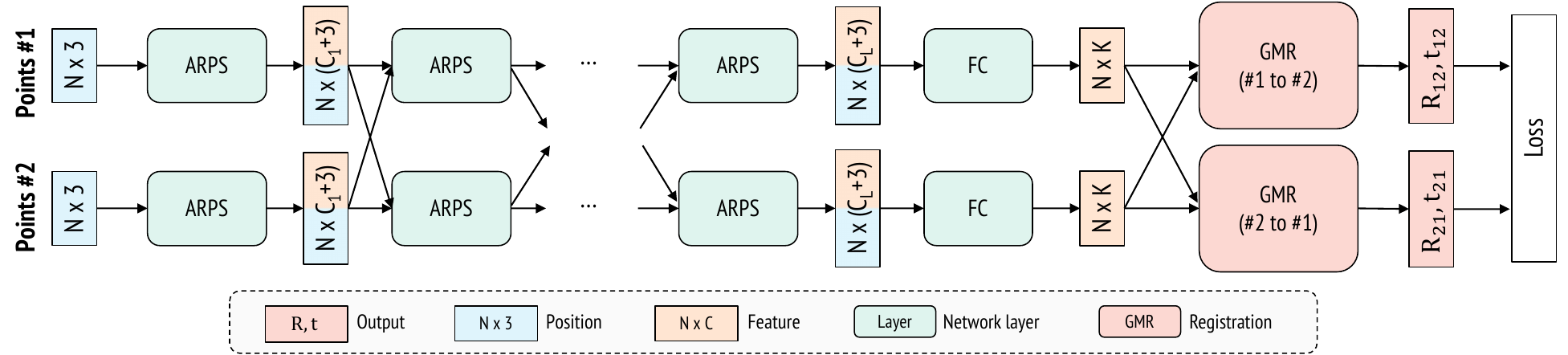}
  \caption{Overall network architecture for GMM-based point set registration using ARPS operations. The network consists of several ARPS layers and a Gaussian mixture registration (GMR) module. In each ARPS layer, a new reference point of each point set is estimated with the features enhanced by attention blocks. Then, the point set is shifted such that the estimated reference point coincides with the origin. Details of ARPS layers are given in \cref{fig:arps-layer}}
  \label{fig:network-architecture}
\end{figure*}

Since the centers of such points of two input point sets will be located at the same position within the global shape, we may use the centers as new reference points. Thus, we translate the source and target point sets such that the estimated reference points approach the origin.
\begin{subequations}
  \label{eq:stepping}
  \begin{align}
    \hat{\mathbf{x}}'_i &= \hat{\mathbf{x}}_i - \frac{\sigma(\alpha)}{H} \sum_{i \in \hat{\mathcal{I}}_H} \hat{\mathbf{x}}_i, \\
    \mathbf{x}'_i       &= \mathbf{x}_i - \frac{\sigma(\alpha)}{H} \sum_{i \in \mathcal{I}_H} \mathbf{x}_i,
  \end{align}
\end{subequations}
where $\alpha \in [-\infty, \infty]$ is a learnable parameter to adjust the step size, and $\sigma$ is the sigmoid function to map $\alpha$ to $[0, 1]$. The adaptive step size stabilizes the training because the centroids estimated in the early layers are less reliable. Reference point shifting allows subsequent ARPS layers to extract more consistent features for partial point sets with an unknown reference point. To supervise the network to obtain a reference point conforming to the centroid of the global shape for each point set, $\hat{\vb{v}}_{\text{gt}}$ and $\vb{v}_{\text{gt}}$, we introduce a centroid loss $\mathcal{L}_{\text{ctr}}$, defined as
\begin{align}
  \mathcal{L}_{\text{ctr}} &= \Bigl\lVert \hat{\mathbf{v}}_{\text{gt}} - \sum_{l} \frac{\sigma(\alpha^l)}{H} \sum_{i \in \hat{\mathcal{I}}_H^l} \hat{\mathbf{x}}_i^l \Bigr\rVert^2 \notag \\
                           &\quad + \Bigl\lVert \mathbf{v}_{\text{gt}} - \sum_{l} \frac{\sigma(\alpha^l)}{H} \sum_{i \in \mathcal{I}_H^l} \!\mathbf{x}_i^l \Bigr\rVert^2,
\end{align}
where variables of the $l$th ARPS layer are denoted by superscript $l$. With this loss, the reference point is guided to be close to the centroid, while it is not necessarily required to be aligned with the centroid.

A notable difference of our method from most attention-based registration methods, such as PRNet~\cite{wang2019prnet}, is that we use the points in the overlap region to find a common reference point for the two input point sets, while most previous studies used them directly as inlier points in the subsequent point-to-point registration module. As pointed out earlier, the use of point-to-point correspondences for partial point set registration may cause inaccurate registration~\cite{huang2020feature}. Thus, we use the overlapping points only to find a reference point, and as we experimentally demonstrate later, it significantly improves the performance of GMM-based registration.

\subsection{Gradual reference point stepping}
\label{ssec:point-stepping}

Even using supervised learning, finding the common centroid is a challenging problem, especially when the poses of two point sets significantly differ. Therefore, we gradually shift input point sets using the estimated new reference points. In this case, a new centroid estimated from point sets with different poses is less reliable. Therefore, we introduce an MLP to estimate the appropriate step size from the differences between mean coordinates and those between mean feature vectors. Then, the output $\alpha := \mathrm{MLP}(\vmu_{\hat{\vb{x}}}^\star - \vmu_{\vb{x}}^\star, \vmu_{\hat{\vb{f}}''}^\star - \vmu_{\vb{f}''}^\star)$ is obtained, where
\begin{subequations}
  \begin{align}
    \vmu_{\hat{\vb{x}}}^\star           &= \frac{1}{H} \sum_{i \in \hat{\mathcal{I}}_H} \hat{\vb{x}}_i, \\
    \vmu_{\hat{\vb{f}}^{''}}^\star \!\! &= \frac{1}{H} \sum_{i \in \hat{\mathcal{I}}_H} \hat{\vb{f}}_i^{''}.
  \end{align}
\end{subequations}
Then, $\alpha$ is activated by the sigmoid function $\sigma$, and the step size $\sigma(\alpha) \in [0, 1]$ is obtained. Then, as we have shown in \cref{eq:stepping}, we translate the source and target point sets such that the estimated reference points approach the origin. To prevent the step size $\sigma(\alpha)$ from being close to 1 and to make the reference points move gradually, we introduce the following penalty for the step size:
\begin{equation}
  \mathcal{L}_{\text{stp}} = \varepsilon \sigma(\alpha)^2,
\end{equation}
where $\varepsilon$ is a small constant set to $\num{1.0e-8}$ by default.

\subsection{Training}
\label{ssec:training}

To train the neural network with ARPS layers, we use four losses: the affine loss $\mathcal{L}_{\text{aff}}$, registration loss $\mathcal{L}_{\text{reg}}$, centroid loss $\mathcal{L}_{\text{ctr}}$, and step size regularizer $\mathcal{L}_{\text{stp}}$. The affine loss and registration loss are the same as those used in the previous study~\cite{huang2022unsupervised}, while the other two are newly introduced in this study. The total loss is simply the sum of these four losses. The proposed network is trained over 400 epochs using the Adam optimizer with learning rate $\gamma=\num{1.0e-4}$ and decay parameters $(\beta_1, \beta_2) = (0.9, 0.999)$. The learning rate is halved when the validation score does not increase for five consecutive epochs. The entire architecture of the proposed network is shown in \cref{fig:network-architecture}.

\section{Experiments}
\label{sec:experiments}

\begin{figure*}[t!]
  \ttabbox[\linewidth]{
    \centering
    \caption{Benchmarking on ModelNet20 with Gaussian noise. GMM-based methods are marked with an asterisk ${}^*$.}
    \label{tab:benchmark-modelnet20}
  }{
    {\tablefontsize
        \begin{tblr}{
          colspec={
              Q[l]
              Q[c,co=1,si={table-format=2.4,round-mode=places,round-precision=4}]
              Q[c,co=1,si={table-format=1.4,round-mode=places,round-precision=4}]
              Q[c,co=1,si={table-format=1.4,round-mode=places,round-precision=4}]
              Q[c,co=1,si={table-format=2.4,round-mode=places,round-precision=4}]
              Q[c,co=1,si={table-format=1.4,round-mode=places,round-precision=4}]
            },
          rowsep=\tablerowsep,
          colsep=5mm,
          hline{2,9,13}={1}{rightpos=-0.2,endpos},
          hline{2,9,13}={2-3}{rightpos=-0.2,leftpos=-0.2,endpos},
          hline{2,9,13}={4-6}{leftpos=-0.2,endpos},
            }
          \toprule
          Method                & {{{MRE \DA}}} & {{{MTE \DA}}} & {{{Recall \UA}}} & {{{RRE \DA}}} & {{{RTE \DA}}} \\
          ICP                   & 18.680341     & 0.171563      & 0.614897         & 6.684228      & 0.101890      \\
          FGR                   & 41.902421     & 0.216454      & 0.602219         & 2.320765      & 0.039875      \\
          RANSAC-10K            & 41.615644     & 0.185725      & 0.663233         & \BB 2.036978  & 0.035220      \\
          Go-ICP                & 21.696097     & 0.098229      & 0.433439         & 4.729202      & 0.058463      \\
          TEASER++              & 74.213131     & 0.416218      & 0.267036         & 7.761676      & 0.073002      \\
          GMMReg${}^{*}$        & 18.503435     & 0.310221      & 0.395404         & 3.089905      & 0.178054      \\
          JRMPC${}^{*}$         & 61.040266     & 0.246900      & 0.423930         & 3.329379      & 0.066090      \\
          PRNet                 & 56.809250     & 0.455531      & 0.061014         & 10.119324     & 0.183860      \\
          RPMNet                & 15.818803     & 0.135855      & 0.630745         & 8.081713      & 0.093143      \\
          DeepGMR${}^{*}$       & 71.955387     & 0.373506      & 0.114897         & 10.078592     & 0.115683      \\
          UGMMReg${}^{*}$       & 36.959432     & 0.285011      & 0.130745         & 10.334652     & 0.128977      \\
          DGMR-ARPS${}^{*}$     & 9.584254      & 0.068956      & 0.829635         & 6.149077      & 0.060828      \\
          DGMR-ARPS+ICP${}^{*}$ & \BB 4.506249  & \BB 0.041849  & \BB 0.915214     & 2.117342      & \BB 0.033567  \\
          UGMM-ARPS${}^{*}$     & 10.798712     & 0.082625      & 0.790016         & 6.961633      & 0.074817      \\
          UGMM-ARPS+ICP${}^{*}$ & 4.703100      & 0.047526      & 0.897781         & 2.208422      & 0.036058      \\
          \bottomrule
        \end{tblr}
      }
  }
  \vspace{1.0\baselineskip}
  \ffigbox[\linewidth]{
    \centering
    \includegraphics[width=\linewidth]{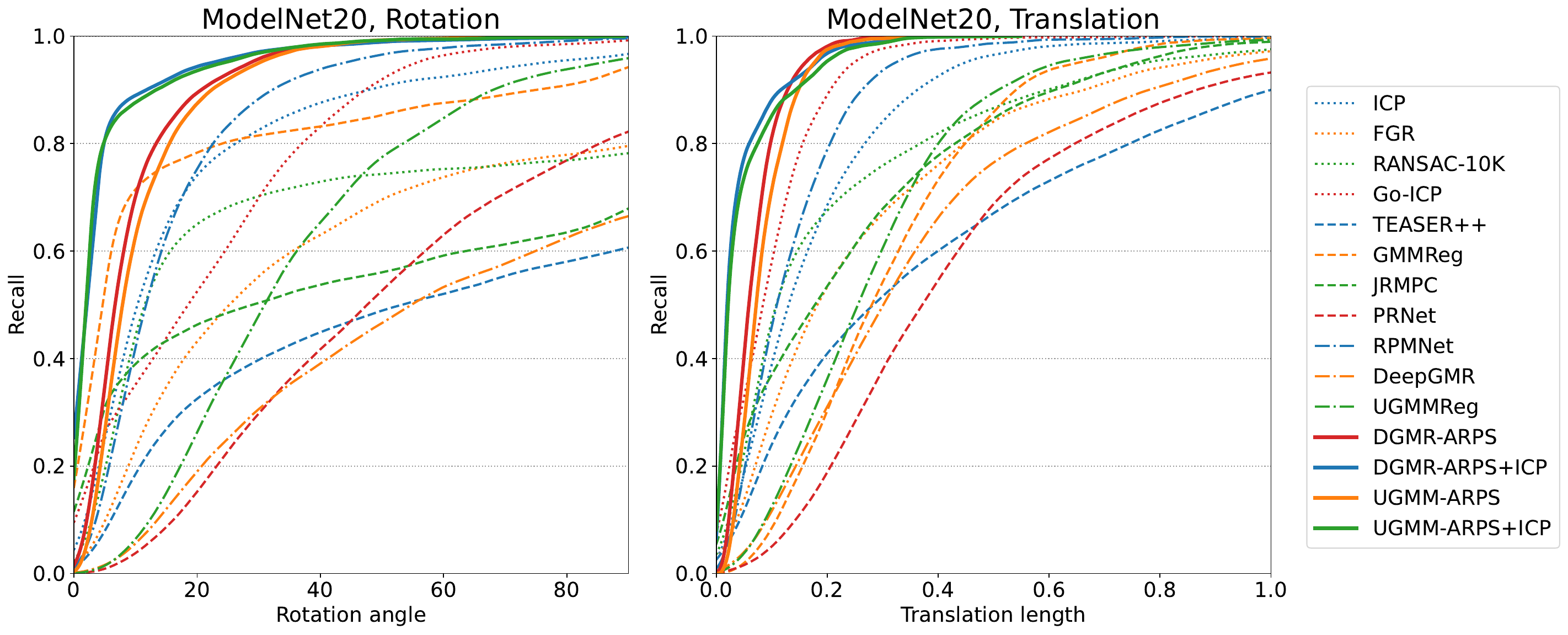}
  }{
    \caption{Recall variation with changing rotation and translation thresholds, for ModelNet20 with Gaussian noise.}
    \label{fig:modelnet20-recall}
  }
\end{figure*}

\begin{figure*}[t!]
  \ttabbox[\linewidth]{
    \centering
    \caption{Benchmarking on ICL-NUIM. GMM-based methods are marked with an asterisk ${}^*$.}
    \label{tab:benchmark-iclnuim}
  }{
    {\tablefontsize
        \begin{tblr}{
          colspec={
              Q[l]
              Q[c,co=1,si={table-format=2.4,round-mode=places,round-precision=4}]
              Q[c,co=1,si={table-format=1.4,round-mode=places,round-precision=4}]
              Q[c,co=1,si={table-format=1.4,round-mode=places,round-precision=4}]
              Q[c,co=1,si={table-format=2.4,round-mode=places,round-precision=4}]
              Q[c,co=1,si={table-format=1.4,round-mode=places,round-precision=4}]
            },
          rowsep=\tablerowsep,
          colsep=5mm,
          hline{2,9,13}={1}{rightpos=-0.2,endpos},
          hline{2,9,13}={2-3}{rightpos=-0.2,leftpos=-0.2,endpos},
          hline{2,9,13}={4-6}{leftpos=-0.2,endpos},
            }
          \toprule
          Method                & {{{MRE \DA}}} & {{{MTE \DA}}} & {{{Recall \UA}}} & {{{RRE \DA}}} & {{{RTE \DA}}} \\
          ICP                   & 39.324726     & 1.035275      & 0.370000         & 4.111030      & 0.136323      \\
          FGR                   & 32.226884     & 1.131394      & 0.290000         & 5.164892      & 0.169245      \\
          RANSAC-10K            & 13.592465     & 0.489297      & 0.725000         & 4.503846      & 0.159716      \\
          Go-ICP                & 23.054986     & 0.311482      & 0.505000         & 4.789978      & 0.142214      \\
          TEASER++              & 82.218297     & 2.449539      & 0.085000         & 6.621835      & 0.166895      \\
          GMMReg${}^{*}$        & 7.131196      & 1.105127      & 0.045000         & 1.847517      & 0.222378      \\
          JRMPC${}^{*}$         & 59.833163     & 0.509504      & 0.475000         & 2.303877      & 0.098589      \\
          PRNet                 & 70.557751     & 2.481688      & 0.010000         & 11.208460     & 0.199467      \\
          RPMNet                & 16.542096     & 0.645177      & 0.285000         & 5.521334      & 0.164931      \\
          DeepGMR${}^{*}$       & 34.203052     & 1.257620      & 0.105000         & 6.662817      & 0.192207      \\
          UGMMReg${}^{*}$       & 25.193895     & 0.962409      & 0.100000         & 5.914902      & 0.194665      \\
          DGMR-ARPS${}^{*}$     & 10.864582     & 0.238078      & 0.620000         & 6.518452      & 0.165839      \\
          DGMR-ARPS+ICP${}^{*}$ & \BB 2.558018  & \BB 0.072464  & \BB 0.900000     & \BB 0.894354  & \BB 0.033060  \\
          UGMM-ARPS${}^{*}$     & 11.870941     & 0.233492      & 0.530000         & 6.679647      & 0.167369      \\
          UGMM-ARPS+ICP${}^{*}$ & 3.484148      & 0.099229      & 0.850000         & 0.954465      & 0.036186      \\
          \bottomrule
        \end{tblr}
      }
  }
  \vspace{1.0\baselineskip}
  \ffigbox[\linewidth]{
    \centering
    \includegraphics[width=\linewidth]{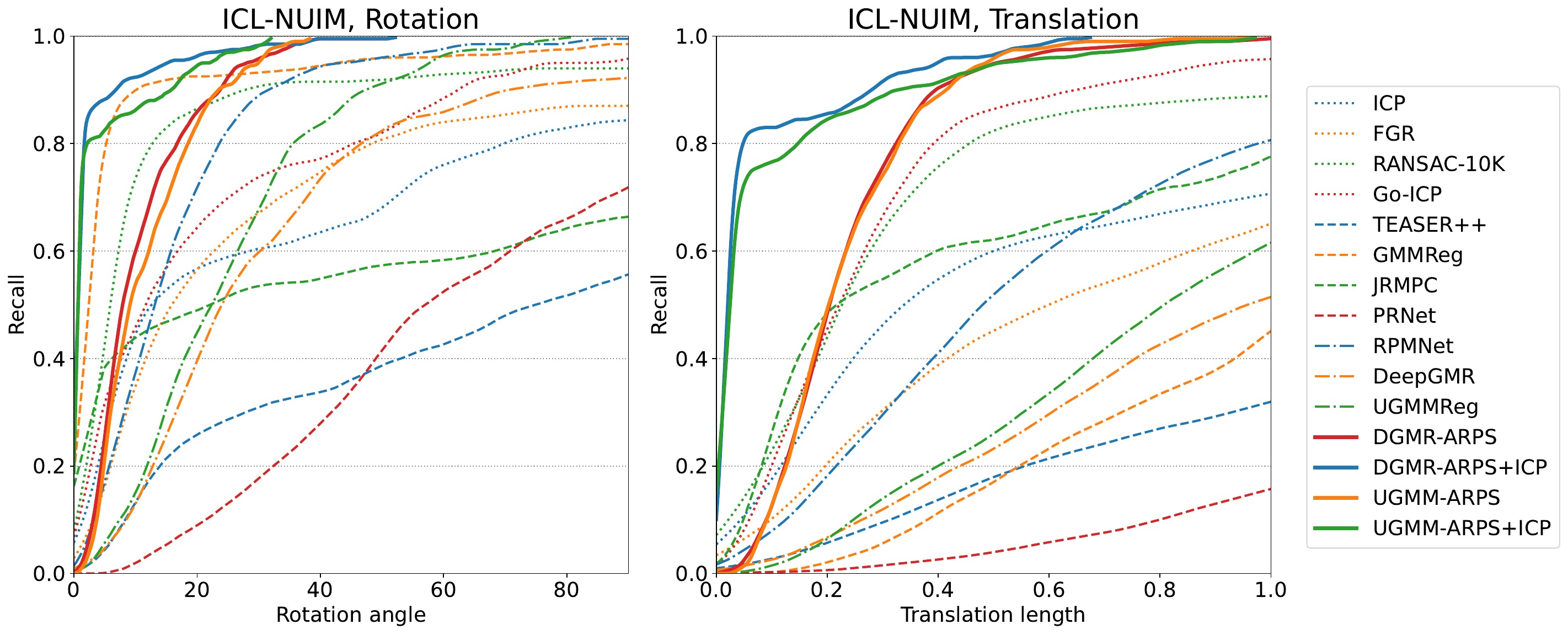}
  }{
    \caption{Recall variation with changing rotation and translation thresholds, for ICL-NUIM.}
    \label{fig:iclnuim-recall}
  }
\end{figure*}

\begin{figure*}[p!]
  \centering
  \begin{tblr}{
      colspec={Q[c]},
      rowsep=-1mm,
    }
    \includegraphics[width=0.98\linewidth]{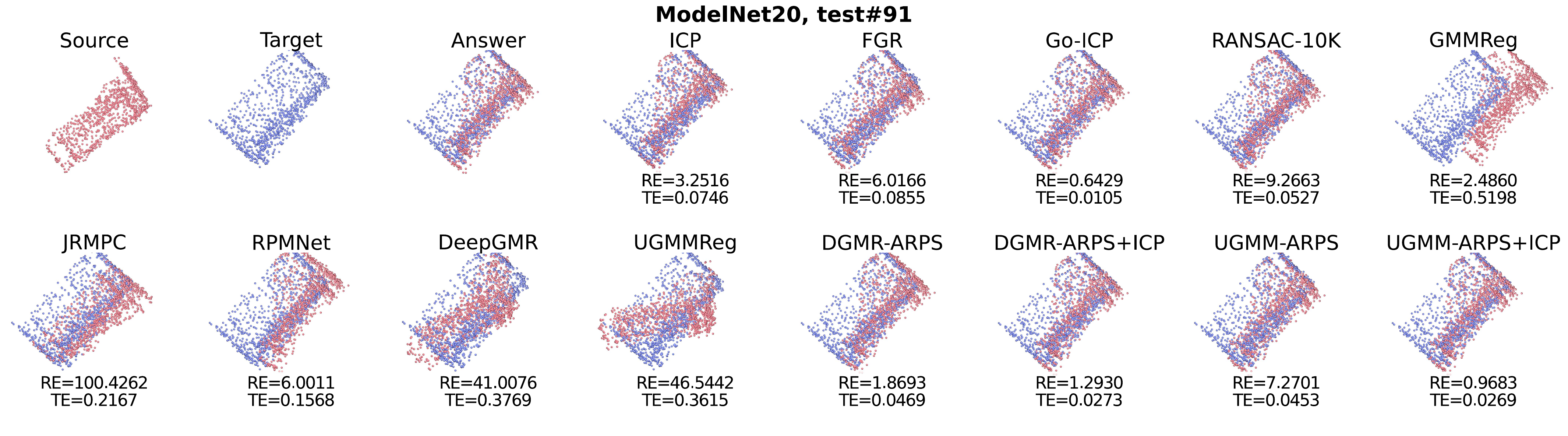} \\
    \includegraphics[width=0.98\linewidth]{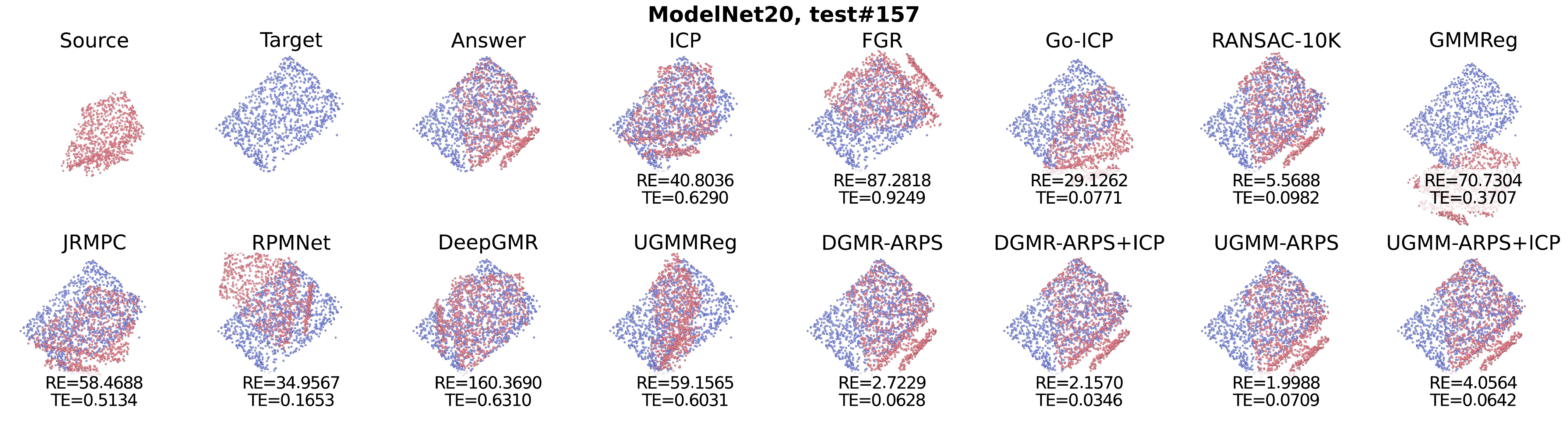} \\
    \includegraphics[width=0.98\linewidth]{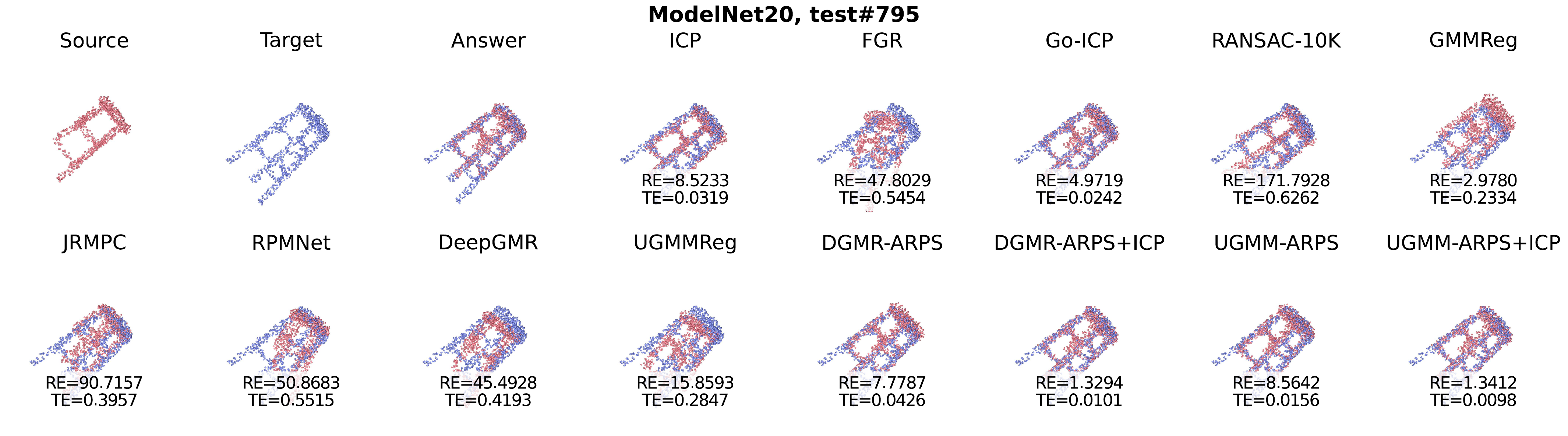} \\
    \includegraphics[width=0.98\linewidth]{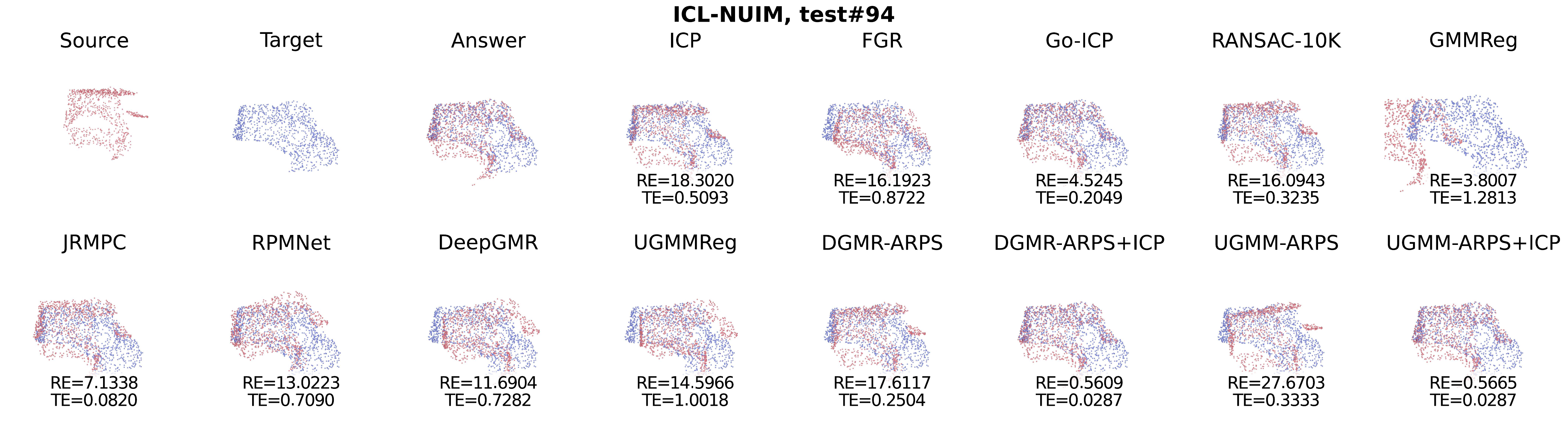} \\
    \includegraphics[width=0.98\linewidth]{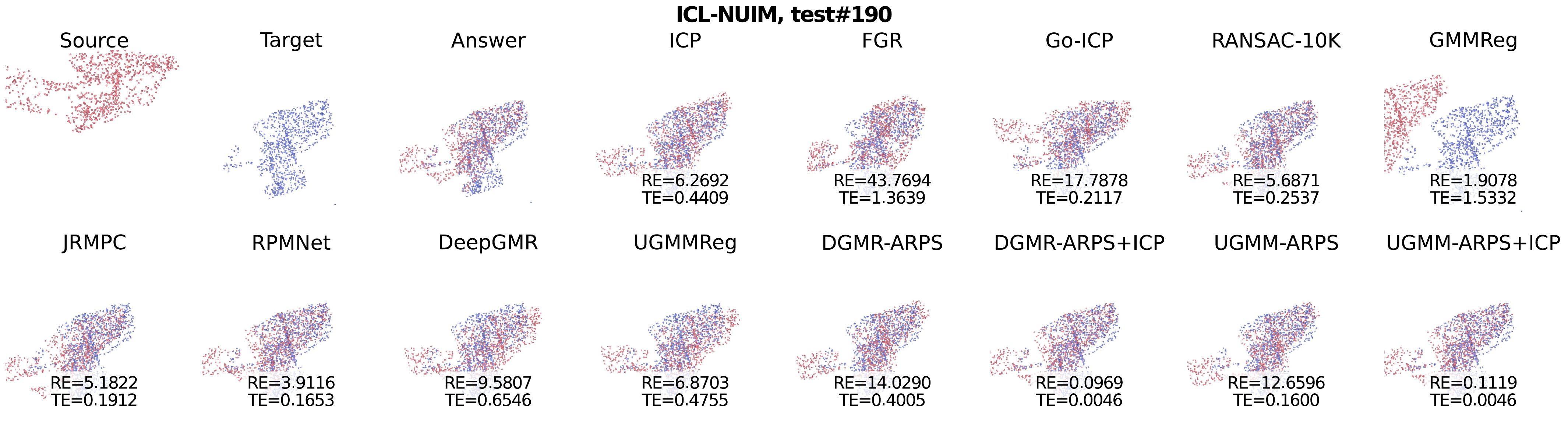} \\
  \end{tblr}
  \caption{Visual comparison of registration results.}
  \label{fig:visual-comparison}
\end{figure*}

\begin{table*}[t!]
  \centering
  \caption{Ablation study}
  \label{tab:ablation-study}
  {\tablefontsize
    \begin{tblr}{
      colspec={
          Q[l,co=1.5]
          Q[l,co=1.5]
          Q[c,co=1,si={table-format=1.4,round-mode=places,round-precision=4}]
          Q[c,co=1,si={table-format=1.4,round-mode=places,round-precision=4}]
          Q[c,co=1,si={table-format=1.4,round-mode=places,round-precision=4}]
          Q[c,co=1,si={table-format=1.4,round-mode=places,round-precision=4}]
          Q[c,co=1,si={table-format=1.4,round-mode=places,round-precision=4}]
        },
      width=0.9\linewidth,
      rowsep=\tablerowsep,
      colsep=5mm,
      hline{2,5,8,11}={1}{rightpos=-0.2,endpos},
      hline{2,5,8,11}={2}{rightpos=-0.2,leftpos=-0.2,endpos},
      hline{2,5,8,11}={3-4}{rightpos=-0.2,leftpos=-0.2,endpos},
      hline{2,5,8,11}={5-7}{leftpos=-0.2,endpos},
        }
      \toprule
      Dataset                     & Method                 & {{{MRE \DA}}} & {{{MTE \DA}}} & {{{Recall \UA}}} & {{{RRE \DA}}} & {{{RTE \DA}}} \\
      \SetCell[r=6]{l} ModelNet20 & ARPS-XYZ               & \BB 9.584254  & \BB 0.068956  & \BB 0.829635     & \BB 6.149077  & \BB 0.060828  \\
                                  & \hspace{1em} No attn.  & 22.794900     & 0.132488      & 0.312995         & 10.192136     & 0.111550      \\
                                  & \hspace{1em} No recen. & 16.079949     & 0.113773      & 0.566561         & 8.775460      & 0.098680      \\
                                  & ARPS-RRI               & 12.939482     & 0.087520      & 0.706815         & 8.178963      & 0.076686      \\
                                  & \hspace{1em} No attn.  & 20.823391     & 0.137555      & 0.367670         & 9.941020      & 0.110824      \\
                                  & \hspace{1em} No recen. & 18.315168     & 0.123446      & 0.475436         & 9.301312      & 0.102810      \\
      \SetCell[r=6]{l} ICL-NUIM   & ARPS-XYZ               & 10.864582     & 0.238078      & 0.620000         & \BB 6.518452  & \BB 0.165839  \\
                                  & \hspace{1em} No attn.  & 20.896991     & 0.434715      & 0.150000         & 8.574272      & 0.207411      \\
                                  & \hspace{1em} No recen. & \BB 10.380407 & \BB 0.235245  & \BB 0.650000     & 6.888075      & 0.171038      \\
                                  & ARPS-RRI               & 14.488299     & 0.344856      & 0.255000         & 7.336984      & 0.190511      \\
                                  & \hspace{1em} No attn.  & 14.899917     & 0.383116      & 0.240000         & 8.205818      & 0.201009      \\
                                  & \hspace{1em} No recen. & 11.904728     & 0.304272      & 0.445000         & 7.910282      & 0.185850      \\
      \bottomrule
    \end{tblr}
  }
\end{table*}

\begin{table*}[t!]
  \centering
  \caption{Effect of the number of ARPS layers. Using four ARPS layers is the baseline and is chosen as the default.}
  \label{tab:num-of-layers}
  {\tablefontsize
    \begin{tblr}{
      colspec={
          Q[l,co=1.5]
          Q[l,co=1.5]
          Q[c,co=1,si={table-format=1.4,round-mode=places,round-precision=4}]
          Q[c,co=1,si={table-format=1.4,round-mode=places,round-precision=4}]
          Q[c,co=1,si={table-format=1.4,round-mode=places,round-precision=4}]
          Q[c,co=1,si={table-format=1.4,round-mode=places,round-precision=4}]
          Q[c,co=1,si={table-format=1.4,round-mode=places,round-precision=4}]
        },
      width=0.9\linewidth,
      rowsep=\tablerowsep,
      colsep=5mm,
      hline{2,5}={1}{rightpos=-0.2,endpos},
      hline{2,5}={2}{rightpos=-0.2,leftpos=-0.2,endpos},
      hline{2,5}={3-4}{rightpos=-0.2,leftpos=-0.2,endpos},
      hline{2,5}={5-7}{leftpos=-0.2,endpos},
        }
      \toprule
      Dataset                     & \#layers     & {{{MRE \DA}}} & {{{MTE \DA}}} & {{{Recall \UA}}} & {{{RRE \DA}}} & {{{RTE \DA}}} \\
      \SetCell[r=3]{l} ModelNet20 & ~~2          & 16.392889     & 0.110782      & 0.565769         & 8.921827      & 0.096249      \\
                                  & ~~4 ($\ast$) & \BB 9.584254  & \BB 0.068956  & \BB 0.829635     & \BB 6.149077  & \BB 0.060828  \\
                                  & ~~6          & 10.906043     & 0.080421      & 0.790016         & 6.941401      & 0.071535      \\
      \SetCell[r=3]{l} ICL-NUIM   & ~~2          & 19.647584     & 0.392470      & 0.185000         & 8.530170      & 0.193406      \\
                                  & ~~4 ($\ast$) & 10.864582     & 0.238078      & 0.620000         & 6.518452      & 0.165839      \\
                                  & ~~6          & \BB 9.803728  & \BB 0.217828  & \BB 0.640000     & \BB 6.581151  & \BB 0.159558  \\
      \bottomrule
    \end{tblr}
  }
\end{table*}

\subsection{Benchmarking on virtual scans}
\label{ssec:benchmarking}

Following the original DeepGMR paper, we compared the performance of our method with those of previous methods using two benchmark datasets, i.e., the ModelNet20~\cite{wu2015shapenets} and augmented ICL-NUIM~\cite{choi2015robust} datasets.

As with the preprocessing in~\cite{wang2019deep,huang2022unsupervised}, we picked a random rotation defined by random Euler angles in $[\ang{-45}, \ang{45}]$ and a random translation defined by a random direction and an amount of translation in $[-0.5, 0.5]$. We also added Gaussian noise $\mathcal{N}(0, 0.01)$ to each point to test the robustness of each method to noise. After preprocessing, we sampled partial point sets with at least \qty{70}{\%} overlap for evaluating partial-to-partial registration performance. As noted previously, the point positions of two input point sets cannot necessarily be aligned even if they belong to the overlapping region. After sampling, each point set consists of \num{1024} points. Point sets from both datasets are normalized to fit within the cube ${[-1, 1]}^3$ by keeping the aspect ratio.

Benchmarking results on these datasets are shown in \cref{tab:benchmark-modelnet20,tab:benchmark-iclnuim}. We compare the proposed method (i.e., DGMR-ARPS) to several traditional approaches, such as standard ICP~\cite{besl1992method}, FGR~\cite{zhou2016fast}, Go-ICP~\cite{yang2013goicp}, feature-based ICP with ten thousand RANSAC iterations (RANSAC-10K), and TEASER++~\cite{yang2021teaser}, and also to GMM-based approaches, GMMReg~\cite{jian2011robust} and JRMPC~\cite{evangelidis2018joint}. Moreover, we also compared it to recent deep-learning-based methods, particularly for partial-to-partial registration: PRNet~\cite{wang2019prnet} and RPMNet~\cite{yew2020rpmnet}, and those using both deep learning and GMMs: DeepGMR and UGMMReg. For a fair comparison, we input only XYZ coordinates to RPMNet, although the original approach also receives other handcrafted features. As in \cref{tab:deepgmr-pretest}, the performance of each method is evaluated using MRE, MTE, recall, RRE, and RTE. In addition, we also show how recall values change with respect to varying rotation and translation thresholds in \cref{fig:modelnet20-recall,fig:iclnuim-recall}. A visual comparison is also shown in \cref{fig:visual-comparison}.

The results indicate that ARPS significantly improves the performances of both DeepGMR and UGMMReg for partial point sets, and thus, our method provides robustness to the inconsistency between a reference point and the centroids of two partial point sets. In addition, performance is further improved using ICP as postprocessing. The proposed method aligns point sets on a distribution basis, and there remains room for improvement in accuracy by performing point-level registration, such as the ICP, as postprocessing.

These results also demonstrate that the proposed method is more robust to differences in local point arrangements than traditional approaches with handcrafted point features. Indeed, the proposed method performs better than RANSAC-10K, the best-performing traditional method. Moreover, the proposed method also performs better than the other deep-learning-based methods.

The weighting scheme of previous methods using an attention block (e.g., PRNet) may somewhat resemble our method. However, by weighting the points to compute a new centroid, our method can more accurately define feature points, resulting in improved correspondence finding. In particular, the attention-based weighting scheme may lead to frequent substitution of points predicted to be in the overlap region. This inconsistency of points in the overlap region during training can cause instability and even failure of the neural network training process. In contrast, our method employs weighting solely for centroid computation. Therefore, our method is more robust to the frequent substitution of points predicted to be in the overlap region.

\begin{table*}[t!]
  \centering
  \caption{Comparison of computational requirements. The ratio of each quantity to that of DeepGMR is shown in parentheses.}
  \label{tab:time-comparison}
  {\tablefontsize
    \begin{tblr}{
      colspec={Q[l]Q[c]Q[c]Q[c]},
      colsep=5mm,
      rowsep=\tablerowsep,
      hline{2}={1}{rightpos=-0.2,endpos},
      hline{2}={2-4}{leftpos=-0.2,endpos},
      }
      \toprule
      Method    & Model size                              & Memory                       & Time                          \\
      DeepGMR   & \makebox[5.5em][r]{1.5M (1.0$\times$)}  & \qty{3.6}{\GB} (1.0$\times$) & \qty{1.73}{\ms} (1.0$\times$) \\
      UGMMReg   & \makebox[5.5em][r]{6.9M (4.5$\times$)}  & \qty{4.6}{\GB} (1.3$\times$) & \qty{4.78}{\ms} (2.8$\times$) \\
      DGMR-ARPS & \makebox[5.5em][r]{13.2M (8.6$\times$)} & \qty{6.2}{\GB} (1.7$\times$) & \qty{8.16}{\ms} (4.7$\times$) \\
      UGMM-ARPS & \makebox[5.5em][r]{13.2M (8.6$\times$)} & \qty{6.2}{\GB} (1.7$\times$) & \qty{8.16}{\ms} (4.7$\times$) \\
      \bottomrule
    \end{tblr}
  }
\end{table*}

\subsection{Ablation study}
\label{ssec:ablation}

We conducted an ablation study using ModelNet20 and ICL-NUIM datasets to reveal which part of the ARPS layer is important for performance improvement. In this experiment, we fed the leading MLPs of the ARPS layers with either XYZ coordinates or RRI features. Moreover, we evaluated the impact of either removing the MHA blocks (both self- and cross-attention) or omitting the recentering operation. The results are shown in \cref{tab:ablation-study}. We see that the MHA block is crucial for achieving high performance, as it significantly enhances results across both datasets. On the other hand, the effect of the recentering operation varies. It improves the performance only for ModelNet20. However, as the performance loss on ICL-NUIM dataset is almost negligible, employing the recentering operation would be a reasonable default for various datasets. Moreover, in agreement with the preliminary experiment in \cref{tab:deepgmr-pretest}, inputting XYZ coordinates rather than RRI features improves the performance. Thus, all the components of the ARPS layer are necessary to improve the performance for partial-to-partial point set registration.

We also examined the effect of changing the number of ARPS layers, as each ARPS layer works to improve the reference point position and to achieve more accurate point set registration. An experiment was conducted with ModelNet20 and ICL-NUIM datasets, with results shown in \cref{tab:num-of-layers}. The performance when using only two ARPS layers is noticeably worse than when using four layers. However, when increasing the number of layers to six from four, the performance does not lead to better performance for both datasets. As using a larger number of ARPS layers requires more time and memory resources, our default choice of using four ARPS layers is a reasonable balance between registration performance and computational demands.

\subsection{Computational demands}
\label{ssec:complexity}

We investigated the increase in memory usage and computation time by introducing our ARPS layers to the original DeepGMR and UGMMReg. \Cref{tab:time-comparison} shows the model size (i.e., number of trainable parameters) for each network, memory resources used during evaluation with mini-batches of 16 point set pairs, and the average computation time required to align a pair of point sets. The quantities shown in this table were measured using a computer equipped with a \qty{3.5}{\GHz} Intel Core i9-11900K CPU with eight cores, an NVIDIA RTX 3080 graphics card, and \qty{64}{\GB} of RAM.

We see that UGMMReg, due to the use of Transformer, has a model size significantly larger than DeepGMR. Because the proposed ARPS layer also has a similar network configuration to that of Transformer, our network has more trainable parameters than UGMMReg, approximately 8.6 times as DeepGMR. The UGMM+ARPS has approximately the same number of trainable parameters as DGMR+ARPS because it uses ARPS layers rather than Transformer.

Despite the increased number of trainable parameters, the memory requirement does not necessarily increase proportionally. As shown in \cref{tab:time-comparison}, the proposed network requires approximately twice as much memory as that of DeepGMR. Although the proposed method takes about five times as long as DeepGMR, the computation time of our method, \qty{8}{\ms}, is sufficiently low in practice. Thus, although the proposed ARPS layer increases the computational load, the increase may be justifiable in practice, given the significant improvement in alignment accuracy provided over the original DeepGMR and UGMMReg.

\begin{figure*}[t!]
  \ttabbox[\linewidth]{
    \centering
    \caption{Benchmarking on KITTI odometry dataset}
    \label{tab:benchmark-kitti}
  }{
    {\tablefontsize
        \begin{tblr}{
          colspec={
              Q[l]
              Q[co=1,c,si={table-format=2.4,round-mode=places,round-precision=4}]
              Q[co=1,c,si={table-format=1.4,round-mode=places,round-precision=4}]
              Q[co=1,c,si={table-format=1.4,round-mode=places,round-precision=4}]
              Q[co=1,c,si={table-format=2.4,round-mode=places,round-precision=4}]
              Q[co=1,c,si={table-format=1.4,round-mode=places,round-precision=4}]
            },
          rowsep=\tablerowsep,
          colsep=5mm,
          hline{2,4}={1}{rightpos=-0.2,endpos},
          hline{2,4}={2-3}{rightpos=-0.2,leftpos=-0.2,endpos},
          hline{2,4}={4-6}{leftpos=-0.2,endpos},
            }
          \toprule
          Method        & {{{MRE \DA}}} & {{{MTE \DA}}} & {{{Recall \UA}}} & {{{RRE \DA}}} & {{{RTE \DA}}} \\
          DeepGMR       & 8.935954      & 1.209843      & 0.094794         & 2.948526      & 0.380937      \\
          UGMMReg       & 12.382903     & 1.266182      & 0.053668         & 3.228287      & 0.372892      \\
          DGMR-ARPS     & 1.801076      & 0.710401      & 0.427446         & 1.569656      & 0.360660      \\
          DGMR-ARPS+ICP & \BB 0.517752  & \BB 0.213193  & \BB 0.915415     & \BB 0.365843  & \BB 0.112774  \\
          UGMM-ARPS     & 2.336193      & 0.792175      & 0.344757         & 1.946372      & 0.369057      \\
          UGMM-ARPS+ICP & 0.568432      & 0.253562      & 0.885664         & 0.365767      & 0.111318      \\
          \bottomrule
        \end{tblr}}
  }
  \vspace{1.0\baselineskip}
  \ffigbox[\linewidth]{
    \centering
    \includegraphics[width=\linewidth]{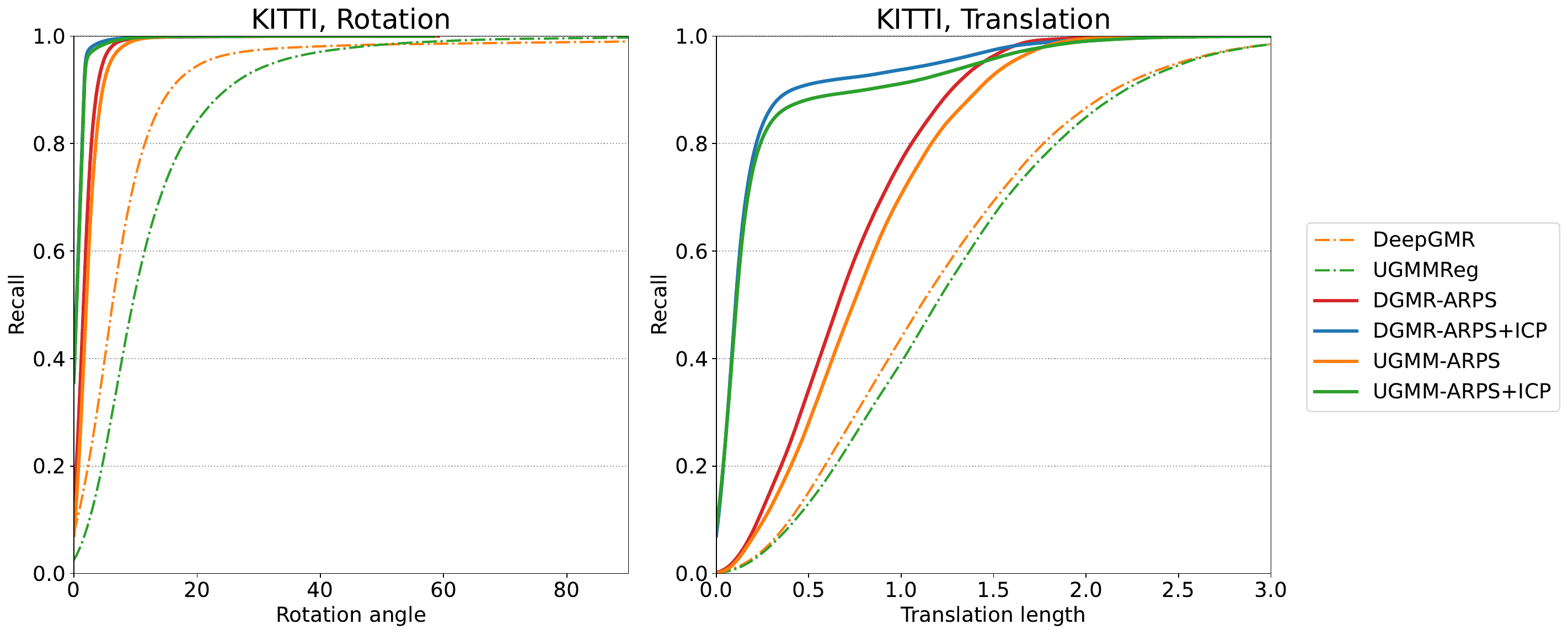}
  }{
    \caption{Recall variation for changing rotation and translation thresholds, for KITTI odometry dataset.}
    \label{fig:kitti-recall}
  }
\end{figure*}

\begin{figure*}[t!]
  \centering
  \begin{tblr}{
    width=\linewidth,
    colspec={X[c] |[dotted] X[c] |[dotted] X[c]},
    colsep=1mm,
    }
    \includegraphics[width=\linewidth]{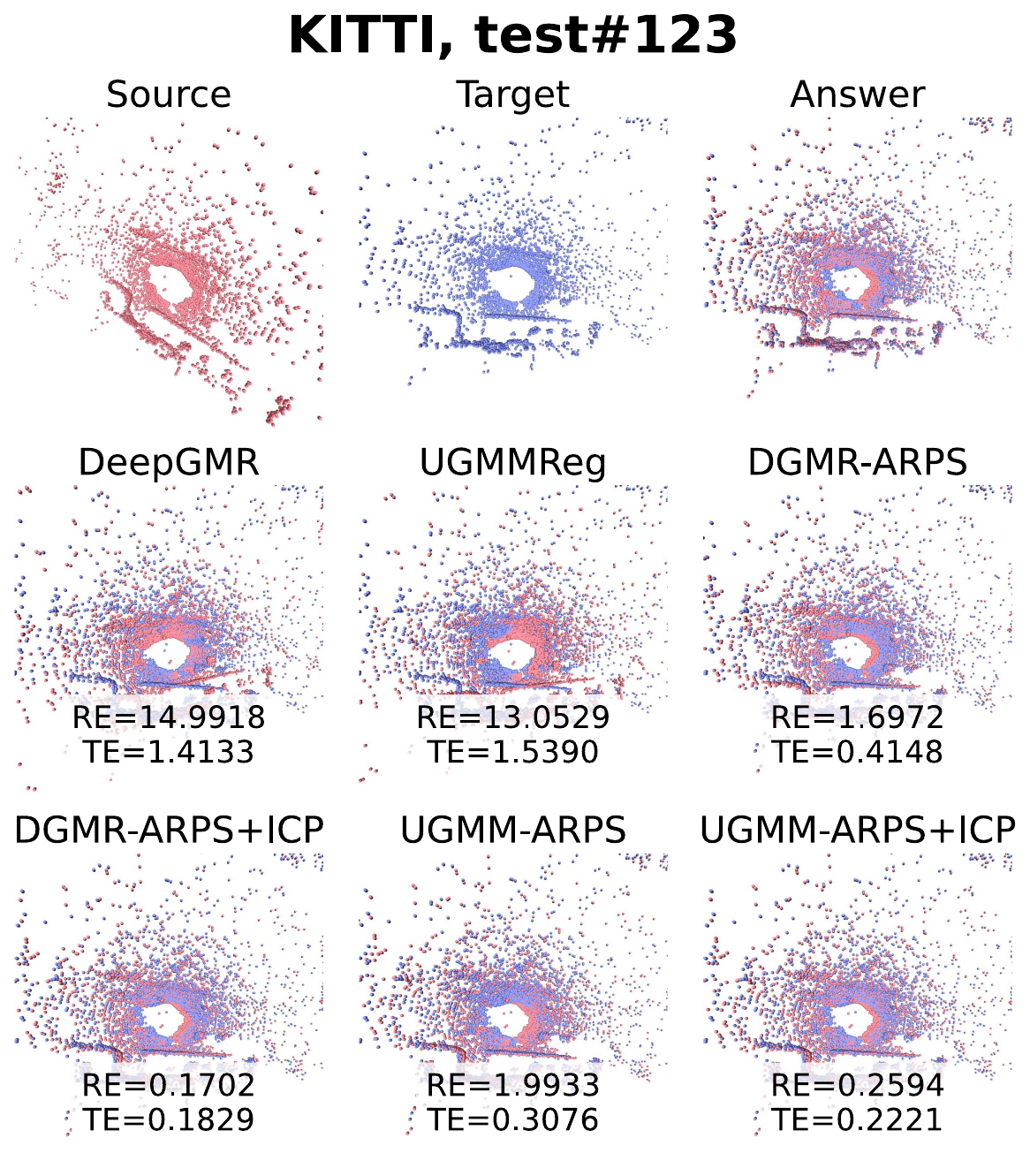} &
    \includegraphics[width=\linewidth]{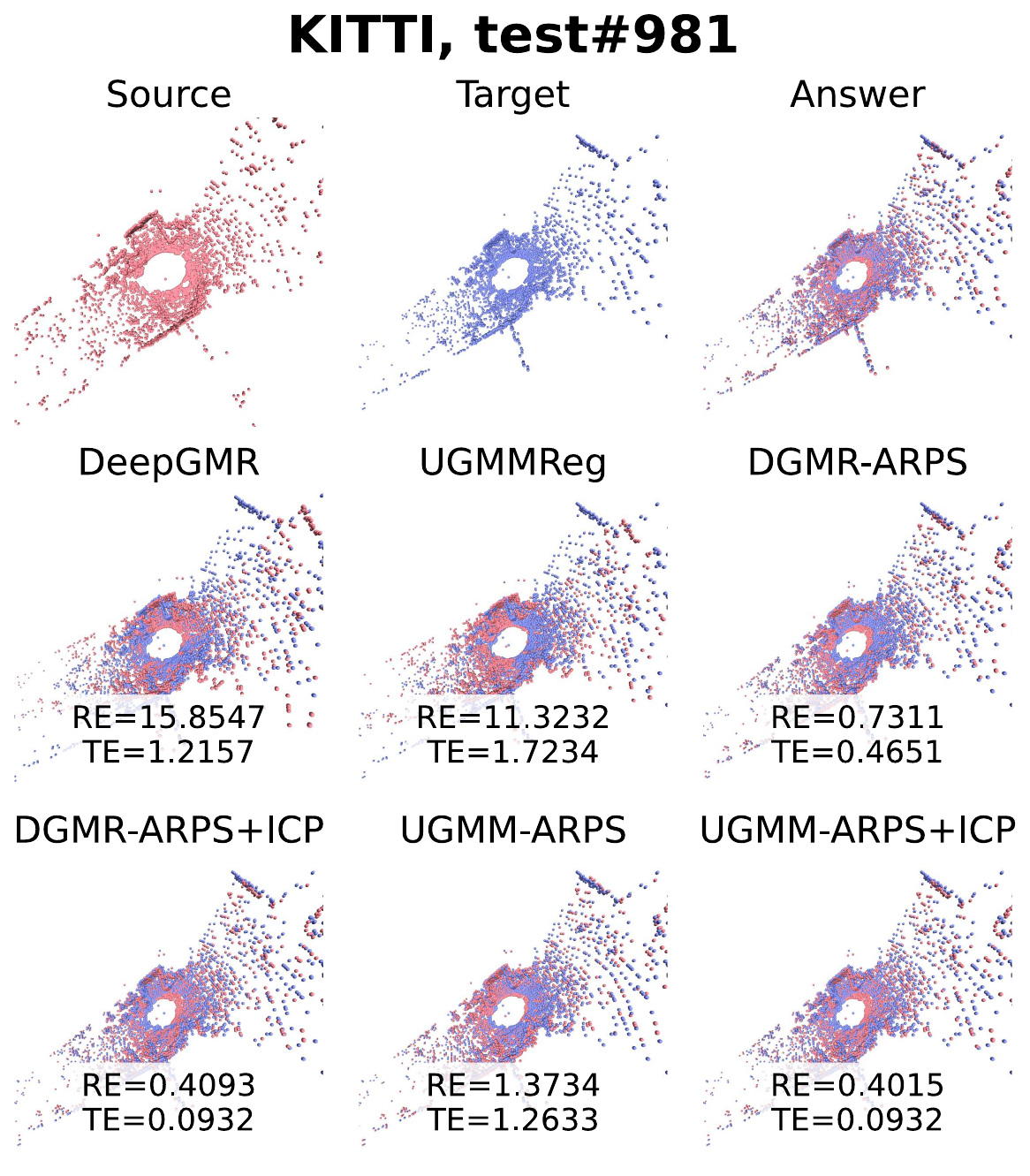} &
    \includegraphics[width=\linewidth]{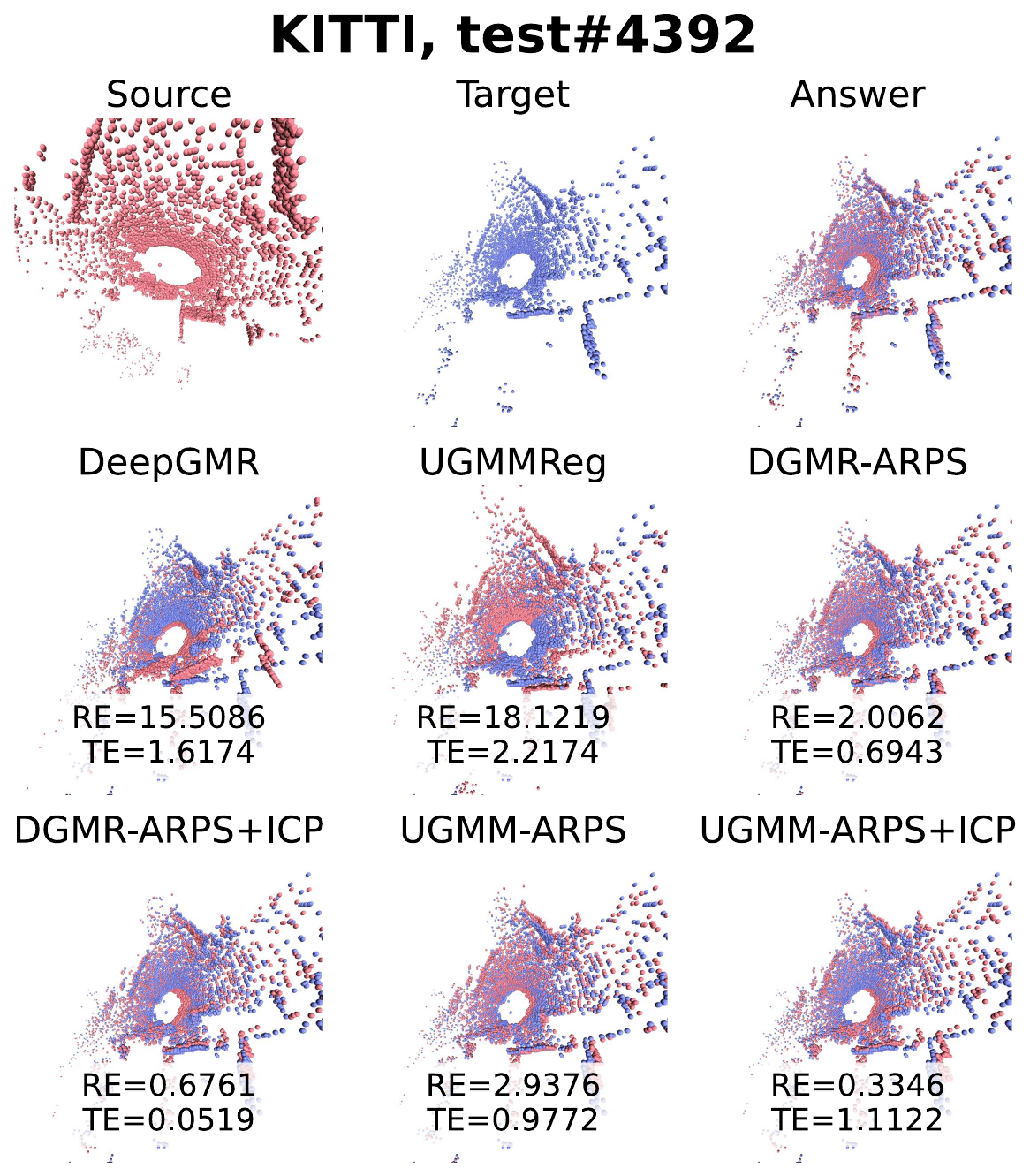}
  \end{tblr}
  \caption{Visual comparison of registration results for KITTI odometry dataset.}
  \label{fig:kitti-visual-comparison}
\end{figure*}

\subsection{Benchmarking on outdoor LiDAR scans}
\label{ssec:bench-real-scans}

We also examined the proposed method with datasets obtained by real LiDAR scans. We used the KITTI odometry dataset~\cite{geiger2012are}, which provides 11 sequences of outdoor LiDAR scans with ground truth camera poses. Following Choy~et~al.~\cite{choy2020deep}, we used scan sequences \#0--\#5 for training, \#6 and \#7 for validation, and \#8--\#10 for testing. To compute the recall score, we used thresholds of \qty{60}{\cm} and \ang{5}.

Quantitative and visual comparisons are shown in \cref{tab:benchmark-kitti}, \cref{fig:kitti-recall}, and \cref{fig:kitti-visual-comparison}. Overall performance of both DeepGMR and UGMMReg has significantly improved by introducing ARPS layers. Moreover, as observed in the previous experiment on ModelNet20 and ICL-NUIM, a refinement step using ICP can further enhance the results. A closer look at \cref{fig:kitti-recall,fig:kitti-visual-comparison} shows that the rotational errors are particularly decreased by the proposed method, which we believe is due to ARPS layers providing a more appropriate reference point and thus gaining rotationally invariant features. Thus, our method also works significantly better than the previous methods for real data given by outdoor LiDAR scanning.

\subsection{Limitations}
\label{ssec:limitation}

While our study has significantly improved GMM-based registration using deep learning, the application of ARPS remains limited to local point set registration. This is largely due to the nature of DeepGMR, which is only effective for duplicated point sets, and has limited applicability as a local registration method for more general cases, e.g., for partial point sets. This limitation arises from the lack of a mechanism to establish optimal correspondences between the Gaussian distributions of two GMMs. Indeed, the proposed method, as well as DeepGMR and UGMMReg, assumes that the Gaussian distributions estimated using the memberships from the neural network have already been sorted such that the distributions with the same index correspond. Addressing this issue would require incorporating an additional process to find soft correspondences between the distributions, akin to recent studies~\cite{wang2019deep,yew2020rpmnet}. However, our preliminary tests with such an approach have not significantly improved the performance of our method. This suggests a need for further exploration and refinement of these approaches.

\section{Conclusion}
\label{sec:conclusion}

In this paper, we introduced a novel approach for GMM-based point set registration using the ARPS layer, which allows a deep neural network to identify a common reference point within partial point sets. By translating the input point sets to align their reference point with the origin, the proposed method allows the neural network to extract consistent point features under rotation and translation. Our approach has made practical the application of deep learning and GMM-based methods to partial-to-partial point set registration, an area previously challenging for such techniques.

Our current framework is optimized for DeepGMR and similar models, emphasizing partial memberships of each point of Gaussian distributions, in contrast to the per-point feature descriptors used in methods like PerfectMatch~\cite{gojcic2019perfect} and Predator~\cite{huang2021predator}. This specialized approach has proven effective for GMM-based approaches, notably enhancing alignment accuracy. As per-point features offer a greater degree of freedom than partial memberships, and as applying our current framework to outperform methods based on such features is not straightforward, we are keen to investigate how our approach can be applied to methods utilizing per-point features. We are also interested in exploring the extension of the current approach to such applications, as traditional GMM-based methods have shown promise in multi-view registration tasks~\cite{evangelidis2018joint}.

\bibliographystyle{CVMbib}
\bibliography{references}

\end{document}